\documentclass[11pt,a4paper]{article}
\usepackage[hyperref]{emnlp2020}

\usepackage{microtype}

\usepackage{times}
\usepackage{latexsym}
\usepackage{graphicx}
\usepackage{hyperref}
\usepackage{amsmath}
\usepackage{amssymb}
\usepackage{array,multirow}

\usepackage{url}
\newcommand{\modelname}[1]{ \textsc{PlotMachines}}
\newcommand{\modelnameshort}[1]{\textsc{PM}}
\newcommand{\singlemem}[1]{\textsc{PM-Single}}
\newcommand{\dualmem}[1]{\textsc{PM-Full}}
\newcommand{\nomem}[1]{\textsc{PM-NoMem}}

\newcommand{\grover}[1]{ \textsc{Grover}}

\newcommand{\taskname}[1]{outline-conditioned story generation}

\aclfinalcopy 

\def\aclpaperid{1687} 

\title{
\textsc{PlotMachines:} \\
Outline-Conditioned Generation with Dynamic Plot State Tracking
}
\author{Hannah Rashkin$^1$,~~~ Asli Celikyilmaz$^2$,~~~ Yejin Choi$^{1,3}$,~~~ Jianfeng Gao$^2$\\
$^1$ Paul G. Allen School of Computer Science \& Engineering, University of Washington\\
$^2$ Microsoft Research, Redmond, WA, USA\\
$^3$ Allen Institute for Artificial Intelligence, Seattle, WA, USA\\
\texttt{\small \{hrashkin,yejin\}@cs.washington.edu}, \texttt{\small \{aslicel,jfgao\}@microsoft.com} \\
}

\begin{document}
\maketitle
\begin{abstract}
We propose the task of \emph{outline-conditioned story generation}: given an outline as a set of phrases that describe key characters and events to appear in a story, the task is to generate a coherent narrative that is consistent with the provided outline. This task is challenging as the input only provides a rough sketch of the plot, and thus, models need to generate a story by interweaving the key points provided in the outline. This requires the model to keep track of the dynamic states of the latent plot, conditioning on the input outline while generating the full story.  
We present \modelname{}, a neural narrative model that learns to transform an outline into a coherent story by tracking the dynamic plot states. In addition, we enrich \modelname{} with high-level discourse structure so that the model can learn different writing styles corresponding to different parts of the narrative. Comprehensive experiments over three fiction and non-fiction datasets demonstrate that 
large-scale language models, such as GPT-2 and \grover{}, despite their impressive generation performance, are not sufficient in generating coherent narratives for the given outline, and dynamic plot state tracking is important for composing narratives with tighter,
more consistent plots.
\end{abstract}

\section{Introduction}

Composing a story requires a complex planning process. First, the writer starts with a rough sketch of what key characters and events the story will contain. Then, as they unfold the story, the writer must keep track of the elaborate plot that weaves together the characters and events in a coherent and consistent narrative. 

\begin{figure}[t!]
\small
    \centering
    \includegraphics[trim={.6cm .5cm 1.7cm .7cm},clip,width=\columnwidth]{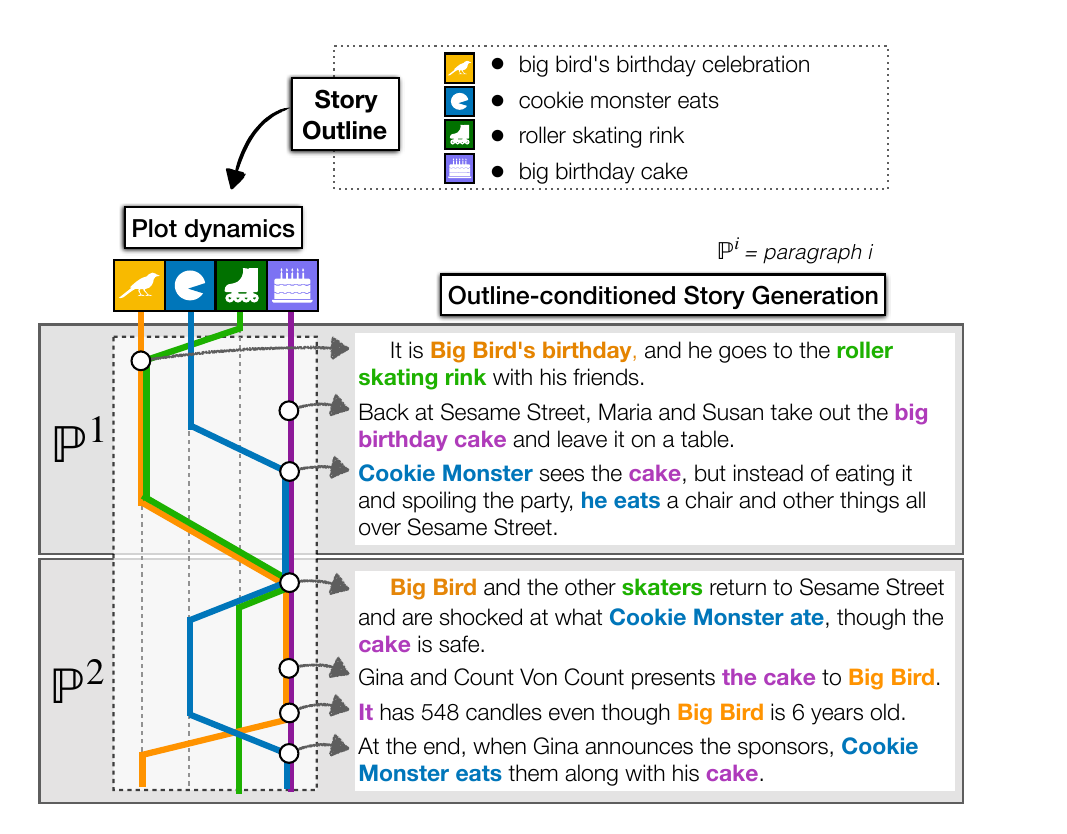}
    \caption{An outline (input) paired with a story (output) from the Wikiplots training set.  
    Plot elements from the outline can appear and reappear non-linearly throughout the plot, as shown in plot dynamics graph. Composing stories from an outline requires keeping track of how outline phrases have been used while writing.
    }
    \label{fig:intro}
\end{figure}

We study this complex storytelling process by formulating it as the task of \emph{outline-conditioned story generation}, illustrated in Figure~\ref{fig:intro}. Given an outline, a set of phrases describing key characters and events to appear in a story, the task is to generate a coherent narrative that is consistent with the provided outline. This task is challenging as the input provides only the rough elements of the plot\footnote{Here, we define plot as the main sequence of events in the story.}.  Thus, the model needs to flesh out how these plot elements will intertwine with each other across different parts of the story. The flowchart in Figure~\ref{fig:intro} demonstrates an example of a latent plot structure: different key phrases from the outline appear and re-appear jointly throughout different sentences and paragraphs.  Notably, the way that outline points are interwoven needs to be determined dynamically based on what's already been composed while also staying true to the original outline and overall narrative structure.

We present \modelname{}, a novel narrative transformer 
that simulates the outline-conditioned generation process described above.\footnote{code available at \url{https://github.com/hrashkin/plotmachines}}
 Our model learns to transform an outline into a multi-paragraph story using dynamic memory blocks that keep track of the implicit plot states computed using the outline and the story generated thus far. We draw inspiration from prior work in dialogue state tracking \citep{DST-1,Lee13,bestdst}, entity tracking \citep{Henaff2016TrackingTW,npn}, and memory networks \citep{endtoendmemnet} for keeping track of plot states.
We also inform our model with high-level narrative structure using discourse labels so that it can learn different styles of writing corresponding to different parts of the narrative (i.e. beginning, middle, and end). \modelname{} is, to the best of our knowledge, the first model designed to generate multi-paragraph stories conditioned on outlines and can be trained end-to-end to learn the latent plot patterns without explicit plot annotations for supervision.

To support research on outline-conditioned generation, we present three datasets, including both fiction and non-fiction domains, where multi-paragraph narratives from existing datasets are paired with automatically constructed outlines using state-of-the-art key phrase extraction. Importantly, our task formulation of outline-conditioned generation is general and can be applied to various forms of grounded language generation. 
Comprehensive experiments on these datasets demonstrate that recently introduced state-of-the-art large-scale language models such as GPT-2 and \grover{} \citep{gpt2,grover}, despite their impressive generation performance, still struggle to generate coherent narratives that are consistent with input outlines.  
Our experiments indicate that dynamic plot state tracking is important for constructing narratives with tighter and more consistent plots compared to competitive baselines.

Our main contributions are: (1) a new task formulation of outline-conditioned story generation, (2) the presentation of three new datasets for this task, (3)  \modelname{}, a novel narrative transformer that learns to transform outlines to full stories with dynamic plot state tracking, and
(4) empirical results demonstrating the limitations of state-of-the-art large-scale
language models and the advantage of \modelname{} compared to competitive baselines.

\begin{table*}[]
    \centering
    \small
    \begin{tabular}{|@{}p{2.5cm}@{}|p{12.5cm}|}
         \hline
        \textcolor{blue}{\textbf{Wikiplots}}
        \  \begin{tabular}{@{}l@{}}
        \ \# stories : 130k \\
        \ avg \# pars : 3.1 \\
        \ data-split : 90/5/5 \\
        \end{tabular}
         & \footnotesize{\textbf{Outline}: $\bullet$ the rocky horror picture show $\bullet$ convention attendees includes servants (...)} {\hskip 4em} 
        \footnotesize{\textbf{Story:} A criminologist narrates the tale of the newly engaged couple, Brad Majors and Janet Weiss, who find themselves lost and with a flat tire on a cold and rainy late November evening, somewhere near Denton in 1974 (...)}
         \\
         \hline
         \textcolor{blue}{\textbf{WritingPrompts}} 
        \begin{tabular}{@{}l@{}}
        \ \# stories : 300k \\
        \ avg \# pars : 5.9 \\
        \ data-split : 90/5/5 \\
        \end{tabular}
          & \footnotesize{\textbf{Outline}: 
          $\bullet$ found something protruding  
$\bullet$ geometric shapes glowing  
$\bullet$ sister kneeling beside 
$\bullet$ dead bodies everywhere 
$\bullet$ darkness overwhelmed 
$\bullet$ firelight flickering 
 (...)} {\hskip 13em} 
         \textbf{Story}: It was dark and Levi was pretty sure he was lying on his back . There was firelight flickering off of what was left of a ceiling . He could hear something but it was muffled . He (...)\\
         \hline
         \textcolor{blue}{\textbf{NYTimes}}
         \begin{tabular}{@{}l@{}}
        \ \# stories : 240k \\
        \ avg \# pars : 15.2 \\
        \ data-split : 90/5/5 \\
        \end{tabular}
         & \footnotesize{\textbf{Outline}: $\bullet$ upcoming annual economic summit meeting $\bullet$ take intermediate steps  (...)} 
         {\hskip 5em} 
         \footnotesize{\textbf{Article:} The long-simmering tensions in Serbia's province of Kosovo turned violent in recent weeks and threaten to ignite a wider war in the Balkans. Only a concerted 
          diplomatic effort by the United States can keep the conflict from escalating. Though he has been attentive to the problem (...)}\\
         \hline
    \end{tabular}
    \caption{Datasets used in the experiments showing the number of stories, the average number of paragraphs per story, and the split of stories across train/dev/test.  We also show an example outline and a short excerpt from a story. We show examples of the full stories in the Supplementary Material.}
    \label{tab:data}
\end{table*}
\section{Outline-Conditioned Generation}
\paragraph{The Task:} 
Our primary goal is to design a task for investigating how story generation models can plan long narrative according to controllable story elements.  
To that end, we introduce the outline-conditioned story generation task, which takes a plot outline as input and produces a long, multi-paragraph story.

In order to be flexible to multiple forms of control that might be required for different downstream tasks, we envision plot outlines to be defined loosely as lists of an arbitrary number of un-ordered plot points that should guide a story being generated.  Plot points could consist of high-level concepts, low-level events, or even detailed sentences. For practical reasons, in this work, we limit the scope of plot points to events and phrases since these can be automatically extracted.  Future work could explore alternate methods of defining plot outlines, perhaps using an event-based planning systems \cite{porteous2009,riedl2010,riedl2010b,fan2019} for generating key points.

More concretely, in this paper, we formulate the outline as a list of un-ordered bullet points which reflect key phrases to be loosely integrated in the output narrative.  These plot outlines are inspired, in part, by previous work in short-form story generation tasks that conditioned on storylines \citep{Peng2018TowardsCS,planandwrite}, which were defined as an ordered list of exactly five single-word points. We extend this concept to long-form story generation by defining a plot outline more flexibly as: an \textit{un-ordered} list of \textit{an arbitrary number of multi-word} plot elements.  An outline also differs from a writing prompt, such as those found in other controllable writing tasks \cite{WritingPrompts}, which are more abstract and often just a starting point for a story.  Unlike a prompt, an outline is a list of concrete points that must appear somewhere in the narrative. 

One challenge of this task is to create stories that have appropriate discourse and narrative flow.  A second challenge is for stories to include the outline in a natural way.  For example, it may be appropriate for certain outline points to be used only later on in the story (e.g. the protagonist dying may be more typically used at the end).

\paragraph{Dataset: Outline to Story:}
\label{datasection}
We construct three datasets for outline-conditioned generation\footnote{Code for replicating data creation available at \url{www.github.com/hrashkin/plotmachines}} by creating novel plot outlines to be used as inputs to generating stories from three existing story datasets. Table~\ref{tab:data} shows statistics and examples from each dataset. 
We focus on fictitious generation, but also include the news domain for generalization.
We build on existing story datasets for the target narratives, which we pair with automatically constructed input outlines as described below:

\textbf{Wikiplots} corpus\footnote{  \url{www.github.com/markriedl/WikiPlots}} consists of plots of TV shows, movies, and books scraped from Wikipedia.  

\textbf{WritingPrompts} \cite{WritingPrompts} is a story generation dataset, collected from the /r/WritingPrompts subreddit $-$ a forum where Reddit users compose short stories inspired by other users’ prompts. 
We use the same train/dev/test split from the original dataset paper.   

\textbf{NYTimes} \cite{nytcorpus} contains news articles rather than fictional stories, unlike the other two datasets.\footnote{Due to concerns over fake news creation, we will not release the model trained on this data.}

\paragraph{Outline Extraction} 
We extract a list of plot outlines from each dataset to use as input using the RAKE (Rapid Automatic Keyword Extraction) algorithm \cite{rake}\footnote{https://pypi.org/project/rake-nltk/}.
RAKE is a domain independent keyword extraction algorithm, which determines key phrases in a document based on the word frequency and co-occurrence statistics. We filtered key-points with overlapping n-grams. This is inspired by similar RAKE-based methods for creating storylines \citep{Peng2018TowardsCS}, but differs in that we extract longer outline points (3-8 words each) with no particular order.

\begin{figure*}
    \centering
    \includegraphics[trim={.2cm .2cm .5cm .7cm},clip,width=\textwidth]{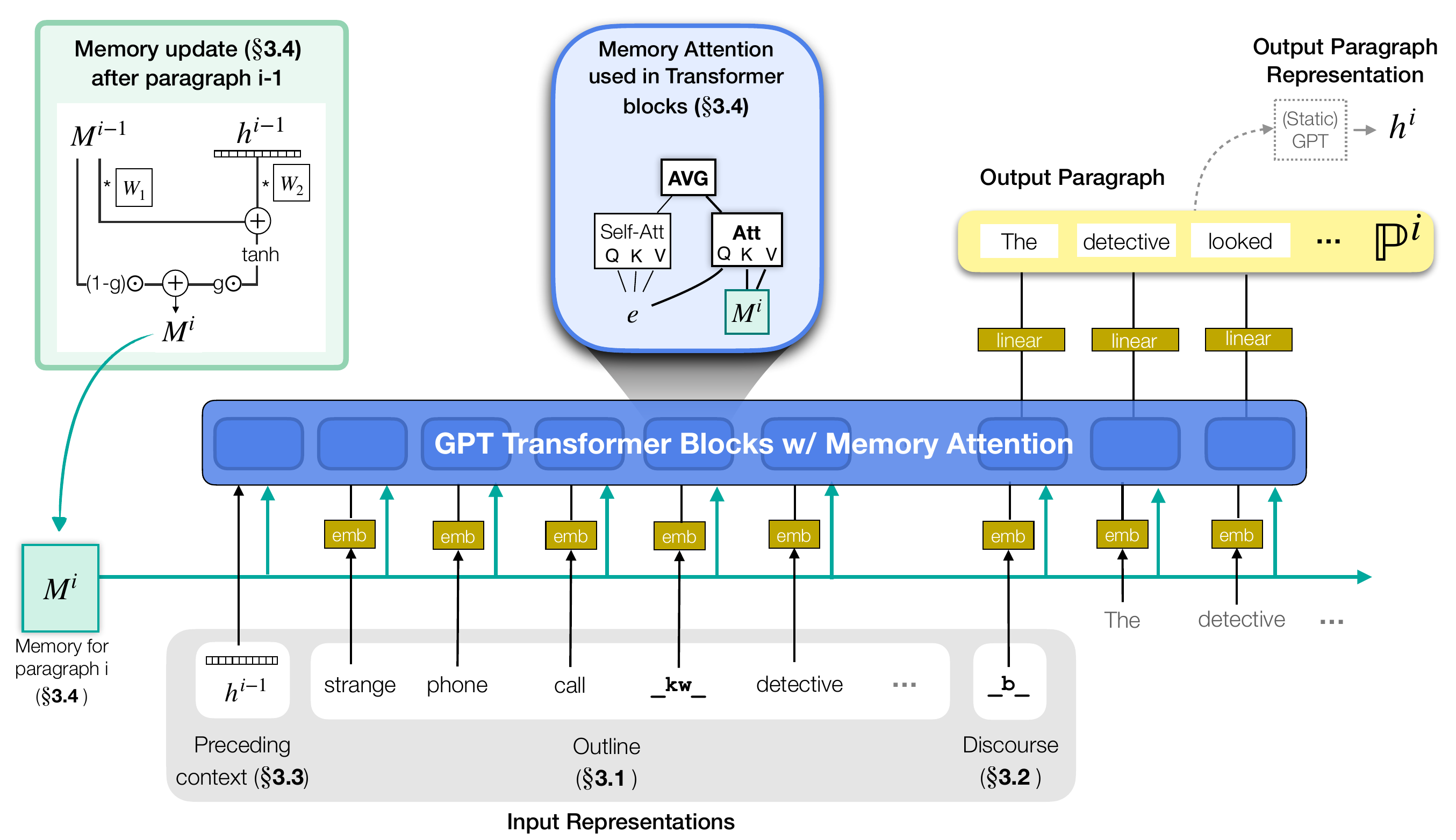}
    
    \vspace{-5pt}
    \caption{\modelname{}: The model generates a paragraph $\mathbb{P}^i$ using the memory ($M^{i-1}$), the previous paragraph representation ($h^{i-1}$), the outline representation ($o$) and discourse representation ($d^i$). First, a gated update mechanism updates the memory using the previous memory and previous paragraph representation.
    Each transformer block includes extra attention over the current memory matrix $M^i$.  The previous paragraph representation, $h^{i-1}$, the outline, and discourse tag (e.g. \texttt{\_b\_}) are also prepended to the generation as an input sequence (gray box).  The output tokens of the generated paragraph are used to compute $h^i$ using a static GPT model. }
    \label{fig:transformers}
    \vspace{-10pt}
\end{figure*}
\section{\modelname{}}
\label{modelsection}
Our approach to this task is to design a model that combines recent success in text generation with transformer-based architectures \citep{vaswanietal} with memory mechanisms that keep track of the plot elements from the outline as they are used in the story.  We also incorporate special discourse features into the modelling to learn a structure over the long multi-paragraph story format.  

We introduce 
\modelname{} (\modelnameshort{}), an end-to-end trainable transformer built on top of the GPT model\footnote{We build on top of GPT, though our approach could be used with most transformer-based LMs.  In experiments, we also look at a version of \modelname{} using GPT-2 \citep{gpt2} as a base.} \cite{GPT}, as shown in Figure~\ref{fig:transformers}. 
Given an outline as input, the model generates paragraphs, recurrently, while updating a memory matrix $M$ that keeps track of plot elements from the outline. This generation framework is motivated by human writing styles, in which each paragraph is a distinct section of related sentences.

At each time step, $i$, \modelname{} generates a new paragraph $\mathbb{P}^i$:
\begin{equation}
    (\mathbb{P}^i, h^i, M^i) = \modelnameshort{}(o, d^i, h^{i-1}, M^{i-1})
\end{equation}
where
$o$ is the outline representation (Sec.~\ref{sec:outlinerep}),
$d^i$ is the discourse representation associated with paragraph $i$ (Sec.~\ref{sec:discourserep}),
$h^{i-1}$ is a vector representation of the preceding story context (Sec.~\ref{sec:neighborrep}), and 
$M^{i-1}$ is the previous memory (Sec.~\ref{sec:memrep}).

\subsection{Outline Representation}
\label{sec:outlinerep}
The plot outline (i.e. the input to the model) is treated as a sequence of tokens, $o$, and used as input for the transformer for each paragraph that is generated.
We use special \texttt{\textbf{\_kw\_}} tokens to delimit each plot point in the outline and end the sequence with a special \texttt{\textbf{\_endkw\_}} token.  We truncate the entire outline to maximum of $n$ tokens.  
For example, an outline containing two plot points (\{`strange phone call', `detective'\}) is turned into the input sequence:
\\
{\small \texttt{strange phone call \textbf{\_kw\_} detective \textbf{\_endkw\_}}}

\subsection{Discourse Representation}
\label{sec:discourserep}
We posit that there are stylistic differences in how the beginning, middle and end of a story are written.  To learn these differences, we introduce $d^i$, discourse information about whether the $i$-th paragraph is an introduction, body, or conclusion paragraph.  We append a special token to the outline representation as part of the input sequence: \texttt{\textbf{\_i\_}}, \texttt{\textbf{\_b\_}}, \texttt{\textbf{\_c\_}} for the introduction, body, and conclusion paragraphs respectively\footnote{We make the simplifying assumption that the first paragraph is an introduction, the last paragraph is the conclusion paragraph, and the other paragraphs are all body paragraphs.}. 

\subsection{Preceding Context Representation}
\label{sec:neighborrep}
With the goal of incorporating previous story context in generating each paragraph, we use $h^{i-1}$, an embedded representation of the previous 
paragraph, which is added to the model input.  More concretely, $h^{i-1}$ is computed as the average embedding of GPT output representations of words from the previous paragraph (using a GPT model that is static, i.e. not finetuned). The $h^{i-1}$ vector is used as an initial input to the transformer architecture, as shown in Figure~\ref{fig:transformers}.

\subsection{Memory Representation}
\label{sec:memrep}
We implement memory to address two key task challenges.  First, we want to keep track of the portions of the outline that have been mentioned. Second, we want to maintain semantic coherence across the entire story. 
To address these two challenges, we implement the memory as consisting of two parts: $K$, a set of vectors keeping track of outline points, and $D$, a matrix that stores a latent topic distribution of what's been written so far.

\noindent \textbf{Notation:} 
We define $d$ as the embedding size of the transformer model and $n$ as the maximum number of tokens in the outline.
Memory is treated as a $\mathbb{R}^{d \times 2n}$ matrix which consists of two smaller matrices stacked together ($M=[K;D]$).
$K$ is a $\mathbb{R}^{d \times n}$ representation of outline points  and $D$ is a $\mathbb{R}^{d \times n}$  representation of the latent document state.  $K$ is initialized with embeddings representing each of the tokens in the outline and $D$ is randomly initialized.  The $j$-th column of memory at the timestep for paragraph $i$ will be denoted $M^{i}_j$.

\noindent \textbf{Updating memory:} 
The memory is updated (top left corner of Fig.~\ref{fig:transformers}) using $h^{i-1}$, the average GPT output representation of the previous paragraph. 
We use update equations based on those in entity-based models such as \citet{Henaff2016TrackingTW}.  We use a gating mechanism, $g$, to allow the model to learn to flexibly control how much each cell in memory is updated, as below:
\begin{align}
\hat{M}^i_j = tanh ( W_1 M^{i-1}_j + W_2 h^{i-1} )\\
g^i_j = sigm(  W_3 M^{i-1}_j +  W_4 h^{i-1})\\
M^{i}_j = (1-g^i_j)\odot M^{i-1}_j + g^i_j\odot \hat{M}^i_j
\end{align}
 where all $W$'s are matrices of dimension $\mathbb{R}^{d \times d}$.
 
\noindent \textbf{Transformer Blocks with Memory:}
Lastly, we must alter the GPT transformer blocks to include the memory in the language modeling.  We alter the attention used within the transformer blocks to contain two parallel attention modules, as shown in Figure~\ref{fig:transformers}. One attention module (on the left in the figure) performs the standard GPT self-attention using transformer inputs to create queries, keys, and values. The other attention module uses transformer input to attend over the memory vectors (i.e., using the memory for creating key and value vectors).  The outputs of both attention modules are averaged\footnote{We experimented with a few other variants of implementing multiple attention mechanisms within the transformer blocks, but found this to be empirically effective.} before performing the remaining transformer block operations.

\subsection{Training and Decoding}
At training time, the model is trained end-to-end on the cross-entropy loss of predicting each paragraph.  Gold representations of previous paragraphs in the story are used to update the memory and compute $h^{i-1}$.  At decoding time, the model must decode a document starting with the first paragraph and use its own predictions to compute $h^{i-1}$ and update the memory.  Additionally, at decoding time, we assume a five paragraph structure (introduction, three body paragraphs, and conclusion) as a pre-set discourse structure to decode from.

\section{Experiments}
We present experiments comparing \modelname{} with competitive baselines and ablations using automatic metrics and human judgements targeting multiple aspects of performance.
In Sec.~\ref{sec:generatedex}, we also include example generations.

\subsection{Experimental Set-up}

\begin{table*}[ht]
\begin{center}
\small
\begin{tabular}{lccc|ccc|ccc}
& \multicolumn{3}{c|}{\textbf{Wikiplots}} & \multicolumn{3}{c|}{\textbf{WritingPrompts}}  & \multicolumn{3}{c}{\textbf{New York Times}}  \\
\hline
Model & R-1 & R-2 & R-L & R-1 & R-2 & R-L & R-1 & R-2 & R-L  \\
\hline
P\&W-Static \citep{planandwrite}&17.0&3.3&13.6& 19.2 & 3.6 & 14.4 &19.3&4.6&15.6\\
Fusion \citep{WritingPrompts}&22.7&6.0&17.4&14.3&1.7&9.6&\textbf{23.2}& \textbf{7.2}&\textbf{18.1}\\
\grover{} \citep{grover}& 19.6 & 5.9 & 12.5 & 23.7 & 5.3 & 17.2 & 20.0 & 5.8 & 14.2\\
\hline
\modelname{}  (GPT) & 20.2 & 5.3 & 16.0  & 30.5 & 5.3 & 25.4 & 21.2 & 5.0 & 15.5 \\
-- base (GPT) \citep{GPT} & 13.2 & 2.0 & 7.9 & 22.1 & 2.7 & 14.3 &  13.9 & 1.6 & 8.3  \\\hline
\modelname{}  (GPT-2) & \textbf{22.8} & \textbf{6.5} & \textbf{17.5}  & \textbf{31.1} & \textbf{6.7} & \textbf{26.1} & {22.1}  & {6.4}  & {16.5}  \\
-- \nomem{} (GPT-2) & 20.5 & 4.9 & 15.5 & 26.6  & 3.7 & 23.5 & 20.0  & 5.4  & 14.4 \\
-- \textsc{PM-NoMem-NoDisc} (GPT-2) & 19.3 & 1.7 & 13.9 & 26.8  & 4.5& 23.2 & 18.4  & 3.4 & 14.2 \\
 -- base (GPT-2) \citep{gpt2} & 18.5 & 3.9 & 13.3 & 26.5 & 4.6 & 20.5 & 19.2 & 4.7 & 13.6 \\
\hline
\end{tabular}
\end{center}
\vskip -0.1in
 \caption{\small ROUGE Results on Wiki, WritingPrompts and NYTimes Datasets. The top block represents the baseline models on story/article generation, while the bottom blocks include ablations of our \modelname{} models.}
\label{table:rouge}
\end{table*}

\paragraph{Baselines:} 
We compare with two models that have been used in related conditional story generation tasks.
First, we train a Fusion model, from the original WritingPrompts dataset paper \citep{WritingPrompts}, using delimited outlines as a single input in place of a prompt.
We also compare with the static storyline-to-story variant of Plan-and-Write (P\&W-Static) from \citet{planandwrite}, which is an LSTM-based model that we train by using the plot outline as delimited input. 

Additionally, given the recent successes in text generation using large pre-trained LM's, we compare with these models, as well.
We finetune the large-scale \grover{} \citep{grover} (equivalent to GPT-2 medium, 345M param)
, which is a transformer-based language model that has been pre-trained for controllable text generation.  
To finetune \grover{}, we give the outline as a delimited form of metadata. 
\grover{} (345M param) has significantly larger capacity than \modelname{} (160M param).
Therefore, for more direct comparison, we also investigate a 460M parameter version of \modelname{} that is built on top of GPT-2 medium \citep{gpt2} instead of GPT.

Unlike our models, the baselines are trained with the traditional generation framework, to generate an entire document conditioned on outlines without generating each paragraph recurrently.

\paragraph{Ablated \modelname{} Models:}
We also show results in Table~\ref{table:rouge} on ablated versions of our model.  First, we use the base GPT and GPT2 models, that are fine-tuned similarly to our model but using only outline inputs (without memory, preceding context, or discourse representations). Second, we investigate the effects of using the preceding context representation but still excluding memory and discourse tokens (\textbf{\textsc{PM-NoMem-NoDisc}}). 
Lastly, we use \textbf{\nomem{}}, a model variant that excludes the memory but uses outline, discourse, and preceding context representations as input.

\paragraph{Details:}
We use the HuggingFace implementations of GPT and GPT-2, and we fine-tune using ADAM.
For generating with our models, we use nucleus sampling with repetition penalties \citep{nucleussampling, CTRL} using $p=90$ and $\theta=1.5$ for GPT and $p=70$ and $\theta=1.4$ for GPT-2 (based on a hyperparameter sweep using grid search with cross-entropy on the dev. data).  We use a minimum sequence length of 100 bpe tokens per paragraph and a maximum sequence length of 400, 922 bpe per paragraph for GPT and GPT-2, respectively. We set $n$, the maximum number of outline tokens and memory dimensions to 100.
We used the settings for the baselines from their respective papers and codebases.

\subsection{Automatic Metrics}

In this section, we evaluate performance using different automatic metrics. We compute ROUGE scores \citep{lin-2004-rouge} and self-BLEU \citep{texygen} following from previous work \citep{shen-etal-2019-towards,texygen} showing that a large ROUGE score together with a low self-BLEU score  can demonstrate a model's ability to generate realistic-looking as well as diverse generations.

\paragraph{Coverage} We compute ROUGE scores \citep{lin-2004-rouge} with respect to the gold stories (Table~\ref{table:rouge}). Results show that the full \modelname{} achieves comparable or higher ROUGE on all three datasets. 
Both \modelname{} variants (using GPT or GPT-2 as a base) achieve improvements over \grover{}, even though \grover{} includes significantly more parameters than the model using GPT.

\begin{table*}[]
    \centering
    \small
    \begin{tabular}{@{}l|c@{}c@{}c@{}c@{}c@{}|c@{}c@{}c@{}c@{}c@{}|c@{}c@{}c@{}c@{}c@{}}
    & \multicolumn{5}{c}{Wikiplots} & \multicolumn{5}{c}{Writing Prompts} & \multicolumn{5}{c}{NY Times}\\
    \hline
        Model & AvgL \ \ & B-2 \ \ & B-3 \ \ & B-4 \ & B-5 \  & AvgL \ \ & B-2 \ \ & B-3 \ \ & B-4 \ & B-5 \ \ & AvgL \ \ & B-2 \ \ & B-3 \ \ & B-4 \ & B-5 \\
         \hline
        Gold Test & 330 & .74 & .50 & .29& .15 & 661 & .82 & .61 & .40 & .25 & 315 & .73 &.50 & .32& .21  \\
        \hline
        P\&W-Static & 352 & .93 & .85 & .75 & .64 &675 & .97& .94& .89& .85 & 352 & .93& .85& .74 & .63\\
        Fusion & 191 & .84 & .71 & .58 &.48&
        197 & .93 & .85 & .75 &.65&
        171&.89 &.80 &.70 &.60\\
        \grover{}& 835 & .72 & .49 & .48 & .37 & 997 & .88 & .72 & .52 & .34 &719 & .79 & .57 & .38 & .25 \\
        GPT & 909 & .77 & .47  & .25 & .11 & 799 & \textbf{.73} & \textbf{.40} & \textbf{.19} & \textbf{.08} & 739 & \textbf{.68} & \textbf{.36} & .27 & \textbf{.08} \\
        GPT-2 & 910 & .60 & .26 & .10 & .03 & 799 & .74 & .41 & \textbf{.19} & \textbf{.08}  & 756 & .69 &  \textbf{.36}& \textbf{.17} &\textbf{ .08} \\
        \hline
        \modelname{} (GPT) & 682 & .77 & .58 & .40 & .27 & 850 & .89 & .81 & .72 & .63 & 537 & .85 & .69 & .53 & .40  \\
        \modelname{} (GPT-2) & 553 & \textbf{.56} & \textbf{.19} & \textbf{.07} & \textbf{.02} & 799 & .83 & .56 & .30 & .14 & 455 & .79 & .57 & .37 & .23 \\
        \hline
    \end{tabular}
    \vskip -0.1in
    \caption{\small Average length of the generated test documents (\textbf{AvgL}) and Self-BLEU n-gram (\textbf{B-n}) scores on 1000 generated story samples from the test sets. We also include the average length and self-BLEU scores of the gold test data. A lower self-BLEU score together with a large ROUGE (see Table~\ref{table:rouge}) score can justify the effectiveness of a model.  We bold the lowest model score in each column; however, we note that sometimes the model self-bleu scores can be lower than in the gold document.}
    \label{tab:genstats}
\end{table*}
\paragraph{Ablations} 
In the bottom block of Table~\ref{table:rouge}, we compare performance of ablated versions of \modelname{}.  First, we compare GPT-2 with \textsc{PM-NoMem-NoDisc}, which differs by including preceding context representations. We observe that \textsc{PM-NoMem-NoDisc} performs slightly better than GPT-2, emphasizing the importance of including context from the previous paragraph.
Second, we investigate the impact of discourse structure representations. We compare \textsc{PM-NoMem-NoDisc}, which omits the discourse token, with \textsc{PM-NoMem}, which uses the discourse token. As shown in Table~\ref{table:rouge}, \textsc{PM-NoMem} generally has higher ROUGE scores than \textsc{PM-NoMem-NoDisc}, indicating that the discourse representation is beneficial to the model.  Lastly, we compare \textsc{PM-NoMem} with  the full \modelname{} to determine the effects of having a memory component. Our full model with  memory  has large ROUGE score improvements over \textsc{PM-NoMem}, underscoring the importance of the plot state tracking. 

\paragraph{Diversity} 
We evaluate the diversity of generated paragraphs from our models using self-BLEU scores \citep{texygen}. 
In Table~\ref{tab:genstats}, we report the self-BLEU scores along with the average length of each generated story. Using all the generated documents from a model, we take one generated document
as hypothesis and the others as reference, and calculate BLEU
score for every generated document, and define the average BLEU
score to be the self-BLEU of the model.  While the Fusion model achieved relatively high ROUGE scores, it has generally worse diversity scores (much higher self-BLEU in Table~\ref{tab:genstats}).  It may be that this model's high ROUGE scores were obtained by producing text that is more repetitive and generic.\footnote{We show an example output from the Fusion model in 
 Figure~\ref{fig:gen:fusion}.}
In contrast, \modelname{} generally achieves good performance on both ROUGE and diversity scores, with self-BLEU scores that are lower than most other models. Notably, they generally have more similar self-BLEU scores to the actual gold stories, indicating that the language diversity is more similar to what humans write.

\begin{figure}
    \centering
    \includegraphics[trim={2.25in 0.65in 0.in 0 },clip,width=\columnwidth]{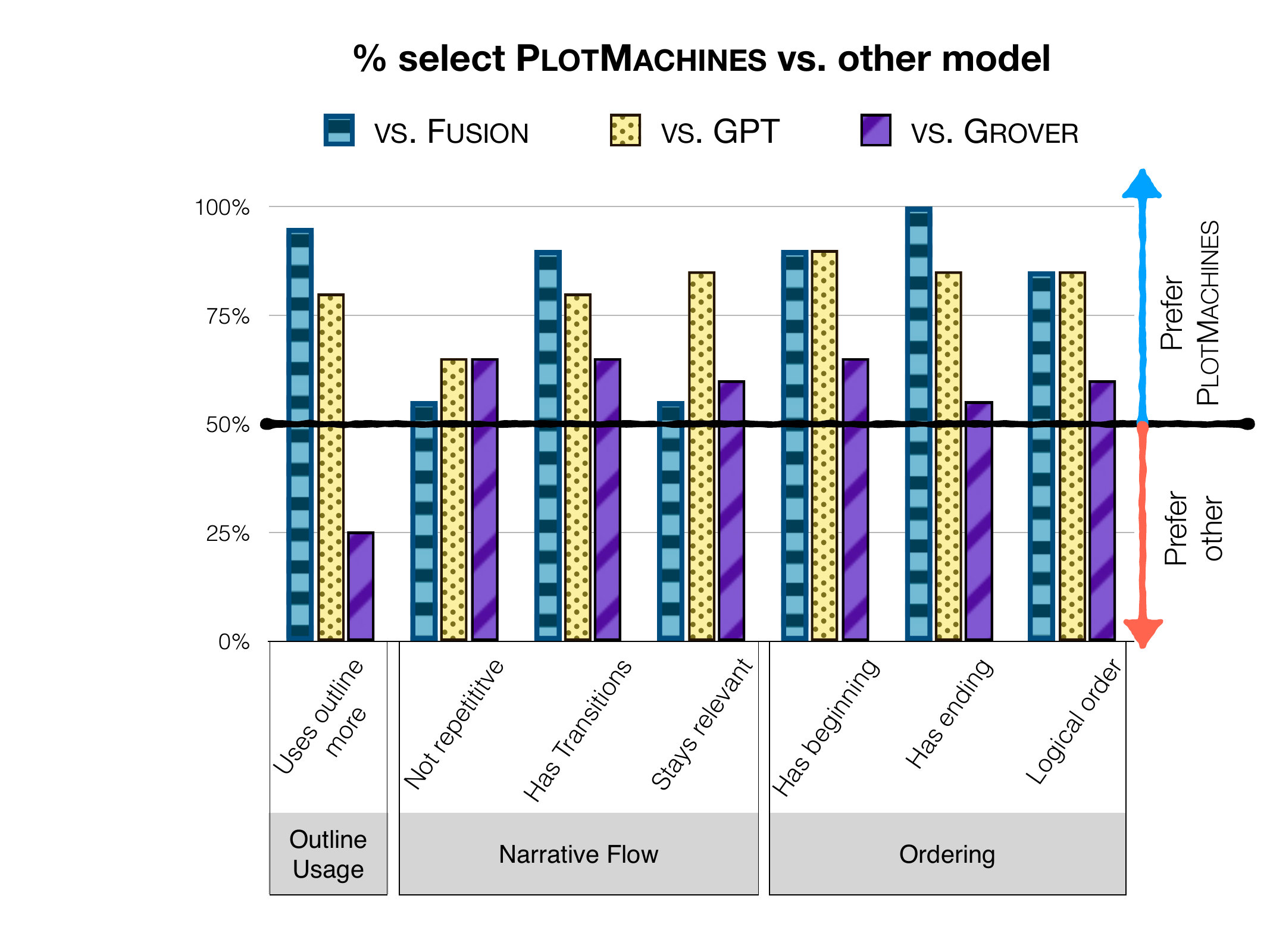}
    \caption{Head-to-head comparsion of \modelname{} vs. three other models for full stories. In the appendix, we report results with standard error metrics (Table~\ref{tab:shortstudy:err}).}
    \label{fig:fullstory}
    \vspace*{-2mm}
\end{figure}

\subsection{Human Evaluations}
Due to the limitations of automatic metrics, we also perform extensive human evaluations.
We conduct human studies to explore how generated stories compare along three dimensions: outline utilization, narrative flow, and ordering.  We ask human raters\footnote{using Amazon Mechanical Turk. We include screenshots from the tasks Appendix ~\ref{sec:screenshots}. In total, over 700 humans participated in all of our studies.} 
to evaluate each single or pair of generations from the Wikiplots test set, collecting 3 responses per story.\footnote{For sampling from Fusion, we restrict to stories or excerpts with two or fewer unk tokens to ensure legibility for workers.}  

To scale up the crowdsourcing work pragmatically, we split evaluations into two studies: one small-scale study evaluating full-length stories, and one large-scale study evaluating single-paragraph excerpts.
In the first study, humans perform head-to-head ratings of 20 randomly sampled stories per pair of models.
In the second study, humans rate story excerpts from 100 randomly sampled outputs per model.

\subsubsection{Full Story Ratings}
\label{sec:human:fullstory}
We give human raters a pair of stories generated from the same outlines and ask them to choose which one is better in different aspects related to outline utilization, narrative flow, and ordering. In Figure~\ref{fig:fullstory}, we show how often \modelname{} (\modelnameshort{}) was selected over the other  models (values above 50\% indicate that \modelnameshort{} was  preferred) using the majority vote for each example.  \modelnameshort{} was selected over base GPT and Fusion in all of the categories, demonstrating that the memory and discourse features are vitally important to improving the base model.  While humans rated \grover{} as using the outline more, \modelnameshort{} 
 is ranked higher in all of the questions about narrative flow and ordering.

\begin{table}[t]
    \centering
    \small
    \begin{tabular}{cc|c}
    \hline
    \multicolumn{3}{c}{Outline Utilization}\\
    \hline
         Model A&  Model B & \% Prefer Model A \\
         \hline
         \multicolumn{3}{c}{Random Paragraph}\\
        \modelname{}&Fusion&	\textbf{80}\% $\pm 4.0$\\
        \modelname{}&GPT&	\textbf{72}\% $\pm 4.5$\\
        \modelname{}&\grover{}&	49\% $\pm 5.0$\\
        \hline
         \multicolumn{3}{c}{Closest Paragraph}\\
        \modelname{}&Fusion&	\textbf{83}\% $\pm $3.8\\
        \modelname{}&GPT&	\textbf{83}\% $\pm 3.8$\\
        \modelname{}&\grover{}&	\textbf{54}\% $\pm 5.0$\\
        \hline
    \end{tabular}
    \vspace*{-1mm}
    \caption{Humans judge which of two paragraphs better utilize the outlines (when shown either random paragraphs or the paragraphs most similar to the outline). }
    \vspace*{-3mm}
    \label{tab:h2h}
\end{table}

\subsubsection{Excerpt Ratings}

\paragraph{Outline Usage}
We give raters two paragraphs each generated by different models and ask them to select which is utilizing the outline better.  
We perform two trials, one with random paragraphs from each story and one with the paragraph from each story that has the most n-gram overlap with the outline (i.e. the closest). In both cases, we compute the majority vote over the three responses and report the percentage of examples where our model is preferred.
Results in Table~\ref{tab:h2h} show that, when looking at single paragraphs, humans tend to choose our \modelnameshort{} as using the outlines in a more natural way, particularly when looking at the ``closest'' paragraph from both models.  Fusion and GPT, in particular, are judged to be utilizing the outline much less than \modelname{}.

\paragraph{Narrative Flow}
In this task, we give raters a generated paragraph (with the previous paragraph as context).  They are asked to rate on a scale from 1 to 5 how much the paragraph: (a) repeats content from the previous paragraph, (b) transitions naturally from the previous paragraph, and (c) stays relevant and on-topic throughout the paragraph.

In the left side of Table~\ref{tab:humaneval}, we show the average ratings of each model.  GPT is the least repetitive between paragraphs but has very low subscores for transitions and relevance. We posit that this behavior is likely due to GPT often generating unrelated content from one paragraph to the next.  
\modelnameshort{} tends to have the highest rated transitions and achieve highest relevancy within paragraphs while being much less repetitive between paragraphs than \grover{} or Fusion.

\begin{table}[]
    \centering
    \small

    \begin{tabular}{c|rrr|r }
         \multirow{2}{*}{Model} & \multicolumn{3}{c|}{Narrative Flow} & \multicolumn{1}{r}{ Order} \\
         \cline{2-5}
          & Rep($\downarrow$) & Tran($\uparrow$) & Rel($\uparrow$)& Acc($\uparrow$) \\
         \hline
         Fusion &  2.61 &2.98&3.36 &\textbf{73}\\
         GPT &  \textbf{1.39}&1.89&2.06 &42\\
         \grover{} &  1.78&3.00&3.29&62\\
         PM & 1.64&\textbf{3.02}&\textbf{3.39}&59\\
         \hline
    \end{tabular}
    \caption{\small Human evaluations of paragraph excerpts from Fusion, GPT, \grover{} and  \modelname{} (PM) outputs.  Narrative flow questions rate the repetitiveness between paragraphs, transitioning, and relevance within paragraphs.  In Table~\ref{tab:humaneval:err} of the appendix, we include standard error metrics.}
    \label{tab:humaneval}
    \vspace*{-2mm}
\end{table}
\paragraph{Ordering}
It's challenging for humans to directly rate the ordering of a story based on a short excerpt. We instead set up a related proxy task: we give raters a pair of consecutive generated paragraphs, presented in a random order, and ask them to attempt to decipher the order.  The intuition is that if the model output is very well-structured then it should be easier for humans to decipher the order. 
We compute the accuracy of the majority vote compared to the actual order in the right side of Table~\ref{tab:humaneval}. 
Accuracy for \modelnameshort{} approaches 60\% accuracy and is much better than the base GPT. \grover{} and Fusion are easiest for humans to re-order (62\%, 73\% respectively).  This result differs slightly from the full story analysis where the humans preferred \modelnameshort{} over \grover{} and Fusion in the ordering-based questions. One possible explanation is that these two models, which decode word-by-word, without an explicit notion of paragraph, may be better at resolving coreference problems between paragraphs.  This may make it easier for humans to re-order short excerpts even though they generally prefer the overall narrative order of \modelnameshort{} due to it having better beginnings, endings, etc. (as indicated in our full story human study).
\begin{figure}
    \centering
    \includegraphics[trim={4.65in 3.8in 1.7in 2.55in},clip,width=\columnwidth]{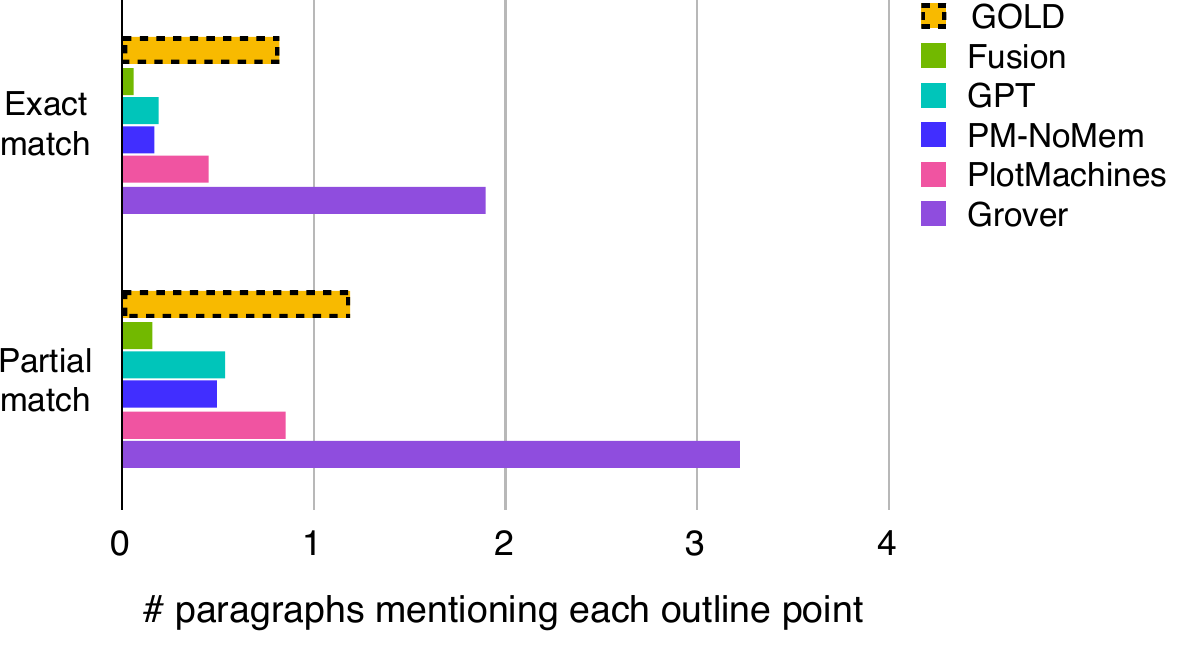}
    \caption{\small Number of paragraphs mentioning each outline point.
    \modelname{} with memory 
    covers points more similarly to the gold story, whereas \grover{} tends to over-repeat outline points (twice as much as the gold reference).}
    \label{fig:ngramoverlap}
\end{figure}

\subsection{N-gram Based Outline Usage Analysis}

We perform an additional quantitative study to further investigate how outline points are used in generated stories.  For fifty stories in the Wikiplots dev. set, we compute how many paragraphs mention each outline point using exact matching or partial matching ($>20\%$ of the n-grams in the outline point also appear in the paragraph). We report the results in Figure~\ref{fig:ngramoverlap}. 

We observe that \grover{} tends to over-repeat outline points (about twice as much as the gold story).  This mirrors our human evaluations that \grover{} is more repetitive. This may also explain why human raters in the full story ratings in Sec.~\ref{sec:human:fullstory} judged \grover{} as using the outline more but having worse narrative flow and order.
Similar observations have been made about pre-trained language models in \citet{See2019} that the models followed story prompts very closely but often copied too much compared to human writing.

In contrast, the Fusion model tends to leave out portions of the outline.  This may reflect the way Fusion was originally designed -- for use with a task using more abstract prompts as input. The GPT and \nomem{} models, while more inclusive than Fusion, are also likely to exclude outline points.  The full \modelnameshort{} model 
is generally more inclusive and more similar to the gold reference than the other models.  The gold story mentions each outline point in around one paragraph on average, indicating that there is an ideal balance between the more conservative coverage achieved by our model and the over-repetitive coverage of \grover{}.

 \subsection{Qualitative Examples}
 In the Appendix,
 (Sec. \ref{sec:generatedex}), 
 we include examples of model outputs on the validation set with annotations for incorporated outline points.  Examples indicate that \grover{} often finishes the story and then starts a new story partway through the document.  This may help explain why \grover{} over-repeats outline points and why humans judge it to be more repetitive and less consistently relevant.
 In contrast, our model adheres more to a beginning-middle-ending structure.

We also look at examples of introduction and conclusion paragraphs generated by \modelname{}, investigating the discourse the model has learned (Table~\ref{tab:introsandconcl}).  The model often starts stories by setting the scene (e.g. ``In the early 1950s, a nuclear weapons testing continues ....'') and tends to write conclusions with a definitive closing action (e.g. ``... the film ends with humperdinck and buttercup riding off into the sunset.'')
\section{Related Work}
\paragraph{State Tracking}
There is a plethora of work in state tracking for dialogue where memory states are updated after each utterance \citep{DST-1,DST-2,Lee13,bestdst}. Similarly, SC-LSTMs \citep{wen-etal-2015-semantically} dynamically updated dialogue act representations as a form of sentence planning in spoken dialogue generation. Memory and entity networks \cite{Henaff2016TrackingTW,endtoendmemnet} and neural checklists \cite{chloe} also used similar methods for tracking entities for other tasks. We adapt these techniques for generating stories while tracking plot state that is updated after each paragraph.
Our method of decoding paragraphs recurrently also draws on existing work in hierarchical decoding \citep{li-etal-2015-hierarchical,shen-etal-2019-towards}, which similarly decodes in multiple levels of abstraction over paragraphs, sentences, and words.

\paragraph{Controllable Story Generation}
There has been a variety of work focusing on generating stories in plot-controllable, plan-driven, or constrained ways (e.g. \citep{riedl2010b,WritingPrompts,Peng2018TowardsCS,Jain2017StoryGF,lebowitz1987planning,ippolito-etal-2019-unsupervised,Perez2001Mexica}).  Similar work in creative generation has conditioned on keywords for poetry generation \cite{Yan2016iPA,Ghazvininejad2016GeneratingTP,Wangetal2016poetry}.  Outline-conditioned generation is complementary to these tasks in that outlines provide more flexibility than very fine-grained srl-based, event-based, or graph-based plans \citep{fan2019,Martin2017EventRF,Harrison2017TowardAS,Li2013StoryGW} and more structured grounding than coarse-grained prompts \citep{WritingPrompts,xuetal} or ending goals \citep{ijcai2019-829}.  Another similar task generates five line stories from five keywords \citep{Peng2018TowardsCS,planandwrite}. We generalize to a similar set-up for long-form narratives. Similar to many recent works in this area, we use seq2seq-based approaches, implemented using  transformers.  We further expand upon the modeling for the challenges specific to our task by using state tracking and applying discourse structure.

\section{Conclusion}
We present \taskname{}, a new task for generating stories from outlines representing key plot elements.  We facilitate training by altering three datasets to include plot outlines as input for long story generation.  In order to keep track of plot elements, we create \modelname{} which generates paragraphs using a high-level discourse structure and a dynamic plot memory keeping track of both the outline and story. Quantitative analysis shows that \modelname{} is effective in composing tighter narratives based on outlines compared to competitive baselines.

\section*{Acknowledgements}
We would like to thank anonymous reveiwers for their insightful feedback.  We also thank Rowan Zellers and Ari Holtzman for their input on finetuning \grover{} and other language models, Maarten Sap for his feedback on human evaluations, and Elizabeth Clark for consulting on baselines and related work. We would also like to thank various members of the MSR AI and UW NLP communities who provided feedback on various other aspects of this work. This research was supported in part by DARPA under the CwC program through the ARO (W911NF-15-1-0543), DARPA under the MCS program through NIWC Pacific (N66001-19-2-4031), and the National Science Foundation Graduate Research Fellowship Program under Grant No. DGE-1256082.

\bibliography{main}
\bibliographystyle{acl_natbib}
\clearpage
\appendix
\section{Supplementary Materials}

\subsection{Examples from Training Datasets}
We show full stories in Tables~\ref{tab:training_data_samples_wiki}-\ref{tab:training_data_samples_nyt} corresponding to the excerpts shown in the Dataset sub-section of Outline-Conditioned Generation in the main text.

\begin{table*}
\centering
\footnotesize
 \begin{tabular}{|p{14cm}|} 
 \hline
\textcolor{blue}{\textbf{Wikiplots Story}} \\ \\
 
 \textbf{Outline}: the rocky horror picture show \textbf{\texttt{\_kw\_}} convention attendees also includes servants riff raff \textbf{\texttt{\_kw\_}} annual transylvanian convention \textbf{\texttt{\_kw\_}} old high school science teache r\textbf{\texttt{\_kw\_}} frank justifies killing eddie \textbf{\texttt{\_kw\_}} enraged rocky gathers frank \textbf{\texttt{\_kw\_}} rainy late november evening \textbf{\texttt{\_kw\_}} dr scott investigates ufos \textbf{\texttt{\_kw\_}} jealous frank kills eddie \textbf{\texttt{\_kw\_}} live cabaret floor show \textbf{\texttt{\_endkw\_}} \\ \\
 \textbf{Article}:
 A criminologist narrates the tale of the newly engaged couple, Brad Majors and Janet Weiss, who find themselves lost and with a flat tire on a cold and rainy late November evening, somewhere near Denton in 1974 . Seeking a telephone, the couple walk to a nearby castle where they discover a group of strange and outlandish people who are holding an Annual Transylvanian Convention . They are soon swept into the world of dr Frank-N-Furter, a self-proclaimed "sweet transvestite from Transsexual, Transylvania" . The ensemble of convention attendees also includes servants Riff Raff, his sister Magenta, and a groupie named Columbia .\\
 
In his lab, Frank claims to have discovered the "secret to life itself" . His creation, Rocky, is brought to life . The ensuing celebration is soon interrupted by Eddie (an ex-delivery boy, both Frank and Columbia's ex-lover, as well as partial brain donor to Rocky) who rides out of a deep freeze on a motorcycle . Eddie then proceeds to seduce Columbia, get the Transylvanians dancing and singing and intrigue Brad and Janet . When Rocky starts dancing and enjoying the performance, a jealous Frank kills Eddie with a pickax . Columbia screams in horror, devastated by Eddie's death . Frank justifies killing Eddie as a "mercy killing" to Rocky and they depart to the bridal suite . \\

Brad and Janet are shown to separate bedrooms, where each is visited and seduced by Frank, who poses as Brad (when visiting Janet) and then as Janet (when visiting Brad) . Janet, upset and emotional, wanders off to look for Brad, who she discovers, via a television monitor, is in bed with Frank . She then discovers Rocky, cowering in his birth tank, hiding from Riff Raff, who has been tormenting him . While tending to his wounds, Janet becomes intimate with Rocky, as Magenta and Columbia watch from their bedroom monitor . \\

After discovering that his creation is missing, Frank returns to the lab with Brad and Riff Raff, where Frank learns that an intruder has entered the building . Brad and Janet's old high school science teacher, dr Everett Scott, has come looking for his nephew, Eddie . Frank suspects that dr Scott investigates UFOs for the government . Upon learning of Brad and Janet's connection to dr Scott, Frank suspects them of working for him; Brad denies any knowledge of it, and dr Scott assures Frank that Brad is totally not involved in UFOs . Frank, dr Scott, Brad, and Riff Raff then discover Janet and Rocky together under the sheets in Rocky's birth tank, upsetting Frank and Brad . Magenta interrupts the reunion by sounding a massive gong and stating that dinner is prepared .\\

Rocky and the guests share an uncomfortable dinner, which they soon realize has been prepared from Eddie's mutilated remains . Janet runs screaming into Rocky's arms, provoking Frank to chase her through the halls . Janet, Brad, dr Scott, Rocky, and Columbia all meet in Frank's lab, where Frank captures them with the Medusa Transducer, transforming them into nude statues . After dressing them in cabaret costume, Frank "unfreezes" them, and they perform a live cabaret floor show, complete with an RKO tower and a swimming pool, with Frank as the leader . Riff Raff and Magenta interrupt the performance, revealing themselves and Frank to be aliens from the planet Transsexual in the galaxy of Transylvania . They stage a coup and announce a plan to return to their home planet . In the process, they kill Columbia and Frank, who has "failed his mission" . An enraged Rocky gathers Frank in his arms, climbs to the top of the tower, and plunges to his death in the pool below . Riff Raff and Magenta release Brad, Janet, and dr Scott, then depart by lifting off in the castle itself . The survivors are then left crawling in the dirt, and the narrator concludes that the human race is equivalent to insects crawling on the planet's surface, "lost in time, and lost in space .  \\
\hline
\end{tabular}
\caption{Example Training Story from Wikiplots Dataset.}
\label{tab:training_data_samples_wiki}
\end{table*}

 \begin{table*}
\centering
\footnotesize
 \begin{tabular}{|p{14cm}|} 
 \hline
 \textcolor{blue}{\textbf{Writing Prompts Story}} \\ \\
 
  \textbf{Outline}: 
 found something protruding \textbf{\texttt{\_kw\_}}
geometric shapes glowing \textbf{\texttt{\_kw\_}}
sister kneeling beside \textbf{\texttt{\_kw\_}}
dead bodies everywhere \textbf{\texttt{\_kw\_}}
running around trying \textbf{\texttt{\_kw\_}}
written anything also \\textbf{\texttt{\_kw\_}}
darkness overwhelmed \textbf{\texttt{\_kw\_}}
firelight flickering \textbf{\texttt{\_kw\_}}
could hear something \textbf{\texttt{\_kw\_}}
definitely screams \textbf{\texttt{\_endkw\_}} \\ \\
 
 \textbf{Article}: It was dark and Levi was pretty sure he was lying on his back . There was firelight flickering off of what was left of a ceiling . He could hear something but it was muffled . He was almost positive it was screaming . When he tried to move he felt an excruciating pain in his left side that caused him to cry out . His hand moved to it instinctively and found something protruding from the wound . It seemed to be a pipe of some kind . He attempted to pull it out but it was covered in blood making it slick . Levi 's ears began ringing and the sounds began to become clearer , it felt quite a bit like some one was driving needles into his eardrums . The sounds he was hearing were definitely screams and not just one person 's , a lot of people were screaming or yelling . There was some one close to him that was crying . He looked in the direction of the tears and seen his sister kneeling beside him , her hands covering her face . `` What happened Laur ? ''. \\
 
 \ \ \ \ \ He was shocked at the sound that barely rasped out from between his lips . His sister 's hands jerked down and she stared down at Levi with a shocked look on her face . `` bu ... I tho ... you were n't breathing ! '' What started as a whisper ended in yell as she threw her self across her brother and began to sob anew . Levi cried out hoarsely but she did n't hear . She just continued to cling to him like a little girl that had just found her lost doll . He put one of his arms around her and scanned the room as much as he could . It looked like a warzone , like something out of one of the many shooters in his gaming collection . `` What the hell ? '' There were dead bodies everywhere , he recognized some of them . There were firefighters and EMT 's running around trying to find survivors in the rubble . Most of the screams were coming from survivors .\\
 
\ \ \ \ \ He seemed to be laying on top of the bottom of a desk , and he was pretty sure the pipe sticking out of his side was a actually one of the legs . Then he spotted it lying about two feet from his right leg , a round section of desk about the size of a softball . On it was a round symbol with geometric shapes glowing with dark red embers and a dull tendril of smoke rising up from it . It all came back to him in rush . He drew that in his notebook . It was second period and his trig teacher had this monotonous voice that could put a crack head to sleep . Laurana caught him doodling and had thrown a pencil at him to scold him silently , which made him jerk as he was connecting the last line on his drawing . Then there was the light and the heat and lastly the dark . Did he do this ? What the hell was going on here ? A voice brought him out of his little flashback . `` Sir . Sir ? Sir ! ''\\ 
 
\ \ \ \ \  He seemed to be laying on top of the bottom of a desk , and he was pretty sure the pipe sticking out of his side was a actually one of the legs . Then he spotted it lying about two feet from his right leg , a round section of desk about the size of a softball . On it was a round symbol with geometric shapes glowing with dark red embers and a dull tendril of smoke rising up from it . It all came back to him in rush . He drew that in his notebook . It was second period and his trig teacher had this monotonous voice that could put a crack head to sleep . Laurana caught him doodling and had thrown a pencil at him to scold him silently , which made him jerk as he was connecting the last line on his drawing . Then there was the light and the heat and lastly the dark . Did he do this ? What the hell was going on here ? A voice brought him out of his little flashback . `` Sir . Sir ? Sir ! ''  \\
 
 \hline
 \end{tabular}
\caption{Example Training Story from WritingPrompts Dataset.}
\label{tab:training_data_samples_wript}
\end{table*}

 \begin{table*}
\centering
\footnotesize
 \begin{tabular}{|p{14cm}|} 
 \hline
 \textcolor{blue}{\textbf{NYT Article}} \\ \\
 
  \textbf{Outline}:
 upcoming annual economic summit meeting \textbf{\texttt{\_kw\_}} take intermediate steps\textbf{\texttt{\_kw\_}} says white house \textbf{\texttt{\_kw\_}} prevent serbian aggression \textbf{\texttt{\_kw\_}} meet boris yeltsin \textbf{\texttt{\_endkw\_}}\\ \\
 
\textbf{Article}:
The long-simmering tensions in Serbia's province of Kosovo turned violent in recent weeks and threaten to ignite a wider war in the Balkans. Only a concerted diplomatic effort by the United States can keep the conflict from escalating. Though he has been attentive to the problem, President Clinton must do more to take the lead with European nations to insure that Kosovo is not left adrift.

\ \ \ \ \ Since Slobodan Milosevic, the Serbian leader, stripped Kosovo of its autonomy in 1989, Kosovo's overwhelmingly Albanian population has engaged in mostly peaceful resistance. It brought them nothing but more repression. In recent months, an Albanian guerrilla army has emerged, targeting mainly Serb policemen. The guerrilla campaign has angered Serbs and given Mr. Milosevic an excuse to bomb villages and carry out indiscriminate attacks. He appears to be trying to push the 1.8 million Albanians out of Kosovo entirely.

\ \ \ \ \ A war in Kosovo, massacres of Albanians or a rush of refugees into Albania and Macedonia could bring those two neighboring countries into the conflict. It might also destabilize the fragile peace in Bosnia and flood Turkey with refugees. Even Turkey and Greece, ancient enemies, might be tempted to intervene to enhance their influence in the Balkans, especially if Macedonia is in chaos.

\ \ \ \ \ International responsibility for dealing with the Kosovo crisis rests primarily with the United States, Britain, France, Italy, Germany and Russia. Acting together as the Contact Group, they are trying to force Mr. Milosevic to accept internationally supervised negotiations with the Albanians. But the group has proved ineffectual because its powers are limited and some members, notably Russia, oppose strong pressure against Serbia. The group has frozen Serbia's assets abroad and this weekend imposed a ban on new foreign investment in Serbia. The sanctions, however, are impossible to enforce among countries outside the Contact Group and difficult even inside it, given Russia's views.     

\ \ \ \ \ When President Clinton meets Boris Yeltsin later this week at the annual economic summit meeting, he should seek more Russian cooperation in pressuring Serbia. He sent a high-level delegation to Belgrade this weekend to say that Serbia will remain isolated if fighting continues. But there is little indication that Mr. Milosevic cares.        

\ \ \ \ \ The White House has not ruled out the use of force to prevent Serbian aggression in Kosovo, but other, intermediate steps should be used before Mr. Clinton considers military action. NATO at this stage can play an important role by increasing its visibility in the region. NATO soldiers ought to be added to a peacekeeping force already based in Macedonia, and a similar group should be stationed in the north of Albania to secure the border and control weapons smuggling. But NATO should also push Mr. Milosevic to accept NATO observers in Kosovo, which he might do if he fears the guerrillas are growing too fast. If Western nations cannot muster a clear and unified message to Mr. Milosevic to restrain his army, he will unleash a new round of ethnic killing in the Balkans. \\
\hline
\end{tabular}
\caption{Example Training Story from New York Times Dataset.}
\label{tab:training_data_samples_nyt}
\end{table*}

\subsection{Human Evaluation Details}
\label{sec:screenshots}
\begin{figure}[h]
    \centering
    \includegraphics[trim={.25cm 0cm 6.9cm 4cm},clip,width=\columnwidth]{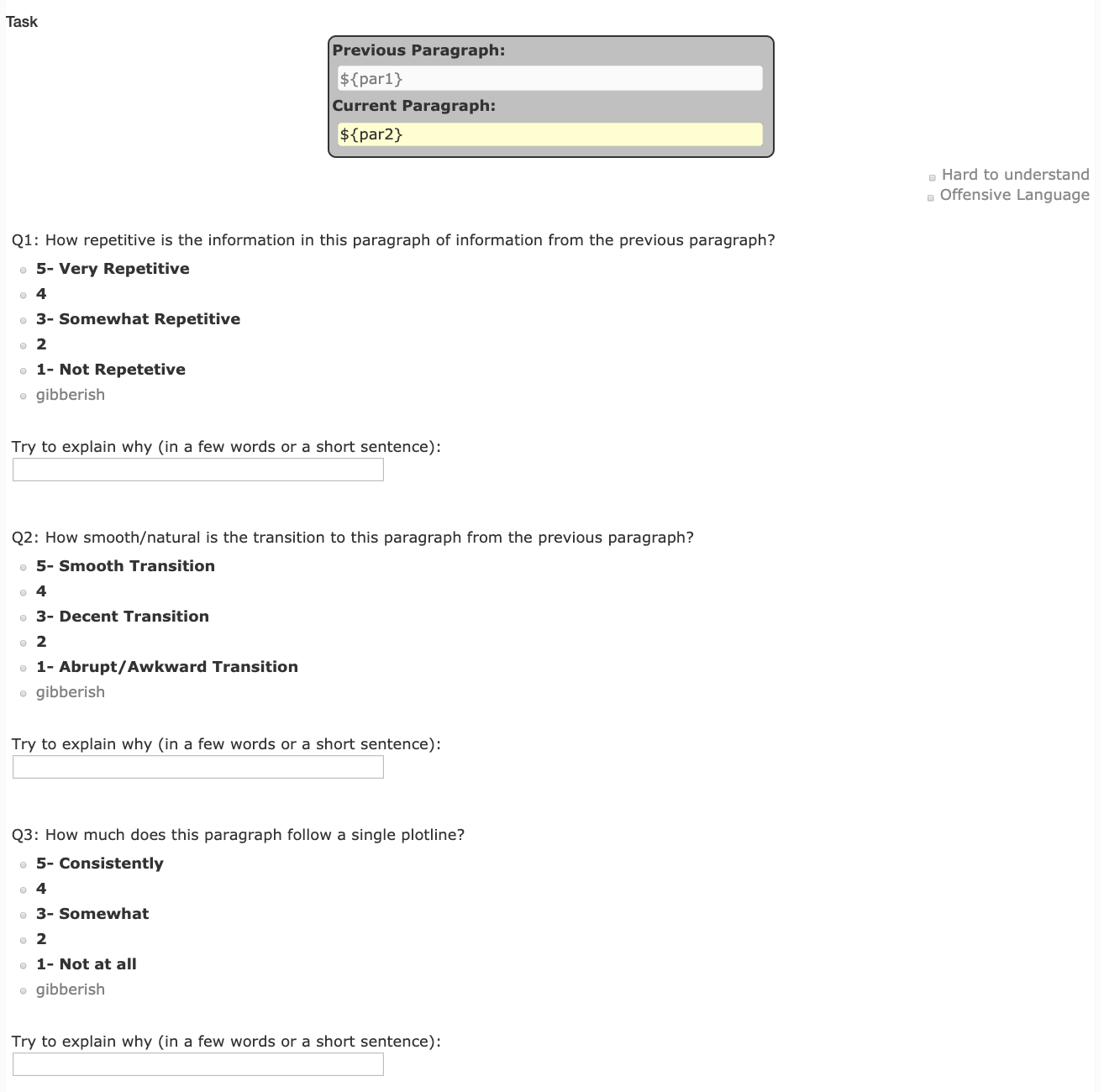}
    \caption{Questionaire for the narrative flow questions about paragraph excerpts.  We pay humans \$1.00 per HIT.}
    \label{fig:narrflowhtml}
\end{figure}

\begin{figure}[h]
    \centering
    \includegraphics[trim={0cm 0cm 11cm 3.5cm},clip,width=\columnwidth]{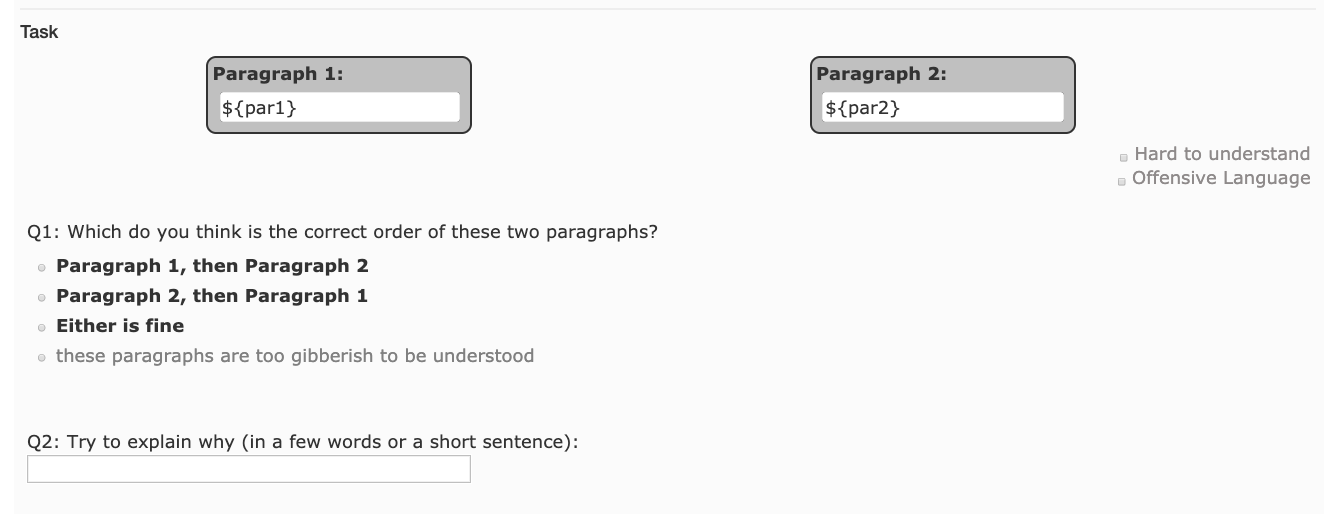}
    \caption{Questionaire for the ordering questions about paragraph excerpts. We pay humans \$1.00 per HIT.}
    \label{fig:orderhtml}
\end{figure}

\begin{figure}
    \centering
    \includegraphics[trim={0cm 0cm 12cm 5cm},clip,width=\columnwidth]{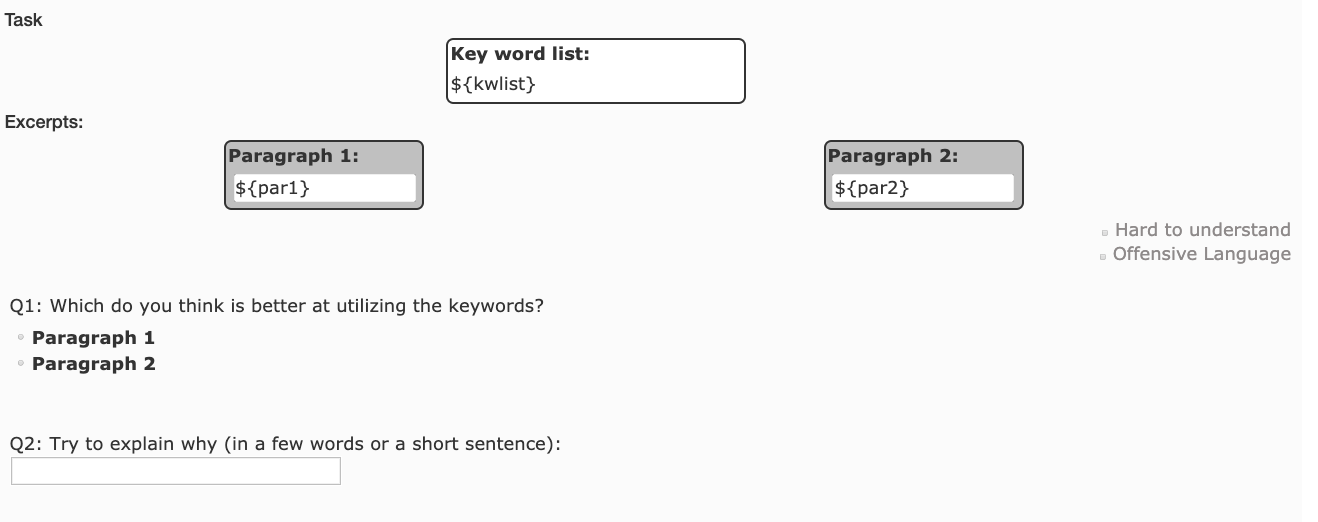}
    \caption{Questionaire for the head-to-head outline usage questions about paragraph excerpts. We pay humans \$1.00 per HIT.}
    \label{fig:outlinehtml}
\end{figure}

\begin{figure}
    \centering
    \includegraphics[trim={0cm 0cm 12cm 5cm},clip,width=\columnwidth]{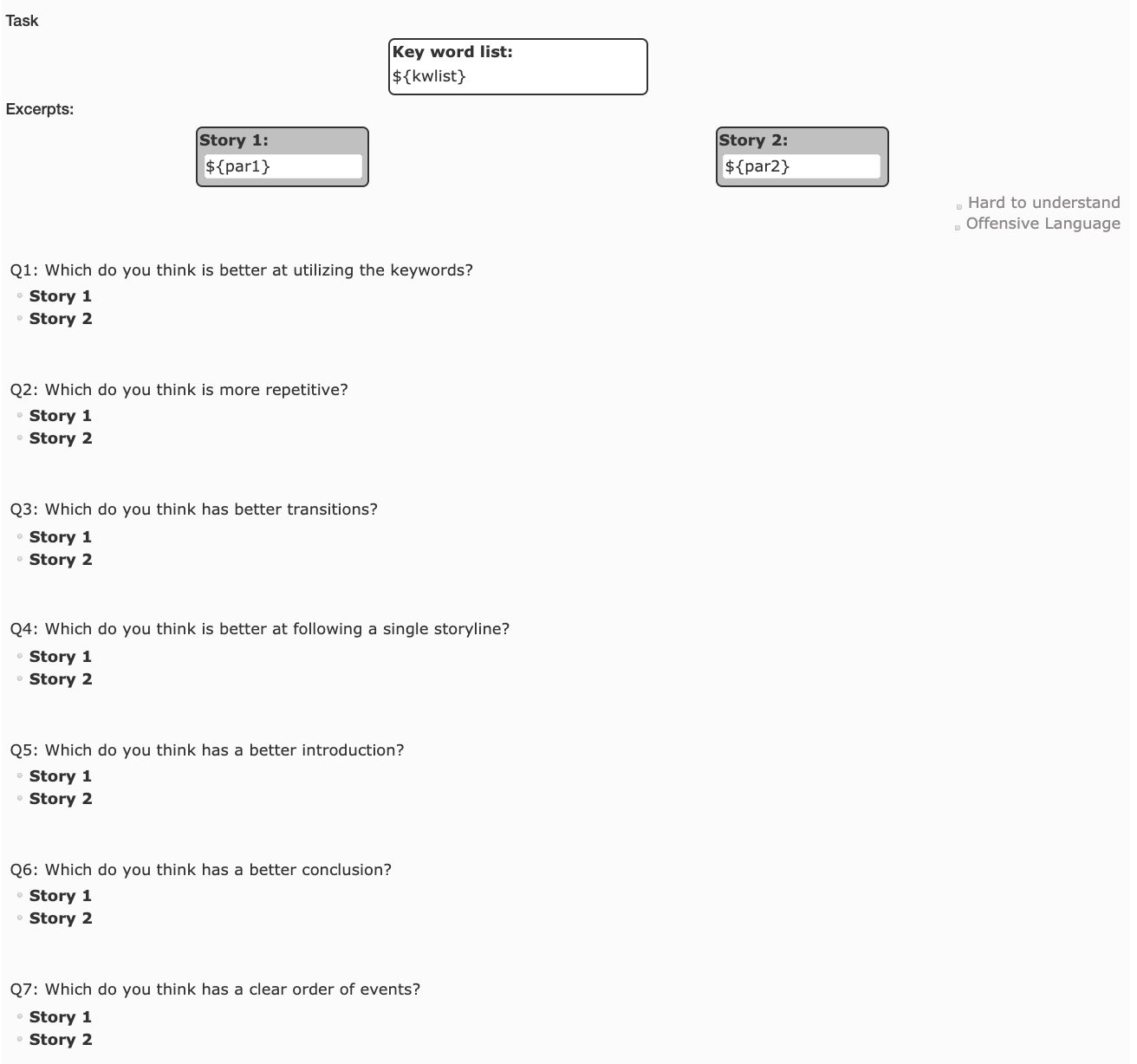}
    \caption{Questionaire for the head-to-head questions about full stories.  We pay humans \$2.00 per HIT.  Note: we reversed the answers to question 2 so that we could show which models were \textit{less} repetitive in Figure~\ref{fig:fullstory}.}
    \label{fig:fullstoryhtml}
\end{figure}

In Figures~\ref{fig:narrflowhtml}-\ref{fig:fullstoryhtml}, we show the questionaires we asked the human raters.  In question 2 of the full story task, we asked about which story was \textit{more} repetitive, but we flip their answers in Figure \ref{fig:fullstory} to show the model that was \textit{less} repetitive in the Figure (i.e. for ease of reading, we made higher better as with the other metrics).


\begin{table*}[]
    \centering
    \small
    \begin{tabular}{cc|rrr }
            &baseline&  \% prefer PM & SEM  & p-val \\\hline
q1-outline  &   Fusion      &95 &4.9    &0.00   \\
q2-repetition   &   Fusion  &55 &11.1   &0.67   \\
q3-transition   &   Fusion  &90 &6.7    &0.00   \\
q4-relevance    &   Fusion  &55 &11.1   &0.67   \\
q5-beginning    &   Fusion  &90 &6.7    &0.00   \\
q6-ending   &   Fusion      &100    &0.0    &0.00   \\
q7-order    &   Fusion      &85 &8.0    &0.00   \\\hline
q1-outline  &   GPT         &80 &8.9    &0.00   \\
q2-repetition   &   GPT     &65 &10.7   &0.19   \\
q3-transition   &   GPT     &80 &8.9    &0.00   \\
q4-relevance    &   GPT     &85 &8.0    &0.00   \\
q5-beginning    &   GPT     &90 &6.7    &0.00   \\
q6-ending   &   GPT         &85 &8.0    &0.00   \\
q7-order    &   GPT         &85 &8.0    &0.00   \\\hline
q1-outline  &   \grover{}       &25 &9.7    &0.02   \\
q2-repetition   &   \grover{}   &65 &10.7   &0.19   \\
q3-transition   &   \grover{}   &65 &10.7   &0.19   \\
q4-relevance    &   \grover{}   &60 &11.0   &0.38   \\
q5-beginning    &   \grover{}   &65 &10.7   &0.19   \\
q6-ending   &   \grover{}       &55 &11.1   &0.67   \\
q7-order    &   \grover{}       &60 &11.0   &0.38   \\
         \hline
    \end{tabular}
    \caption{\small Small-scale human study: H2H comparison of \modelname{} (PM) with baseline output for 20 full stories. SEM is the standard error of the mean. The p-value is a t-test comparing to 50\% (no preference between outputs). Although this is a small-scale study, the preference for \modelnameshort{} is significant in many of the comparisons to Fusion and GPT2. Overall, there is a general trend towards \modelnameshort{} being preferred in all cases except for the comparison of outline utilization with \grover{}.}
    \label{tab:shortstudy:err}
\end{table*}

\begin{table*}[]
    \centering
    \small
    \begin{tabular}{c|rrr|r }
         \multirow{2}{*}{Model} & \multicolumn{3}{c|}{Narrative Flow} & \multicolumn{1}{r}{ Order} \\
         \cline{2-5}
          & Rep($\downarrow$) & Tran($\uparrow$) & Rel($\uparrow$)& Acc($\uparrow$) \\
         \hline
         Fusion &   2.61$\pm0.09$ &   2.98$\pm0.08$&  3.36 $\pm0.08$&    \textbf{73}$\pm4.4$\\
         GPT &  \textbf{1.39}$\pm0.06$&  1.89$\pm0.09$&   2.06$\pm0.10$ &    42$\pm4.9$\\
         \grover{} &  1.78$\pm0.08$& 3.00$\pm0.11$&   3.29$\pm0.11$& 62$\pm4.9$\\
         PM & 1.64$\pm0.07$& \textbf{3.02}$\pm0.10$&   \textbf{3.39}$\pm0.10$&  59$\pm4.9$\\
         \hline
    \end{tabular}
    \caption{\small Extended results with standard error of the mean for human evaluations of paragraph excerpts from Fusion, GPT, \grover{} and  \modelname{} (PM) outputs.  Narrative flow questions rate the repetitiveness between paragraphs, transitioning, and relevance within paragraphs.}
    \label{tab:humaneval:err}
\end{table*}

\subsection{Qualitative Examples}
\label{sec:generatedex}
In this section, we include examples of model outputs on the validation set with annotations for incorporated outline points.

\begin{figure*}
\centering
\includegraphics[page=3,trim={0 11.in 0 0in}, clip,height=9.in]{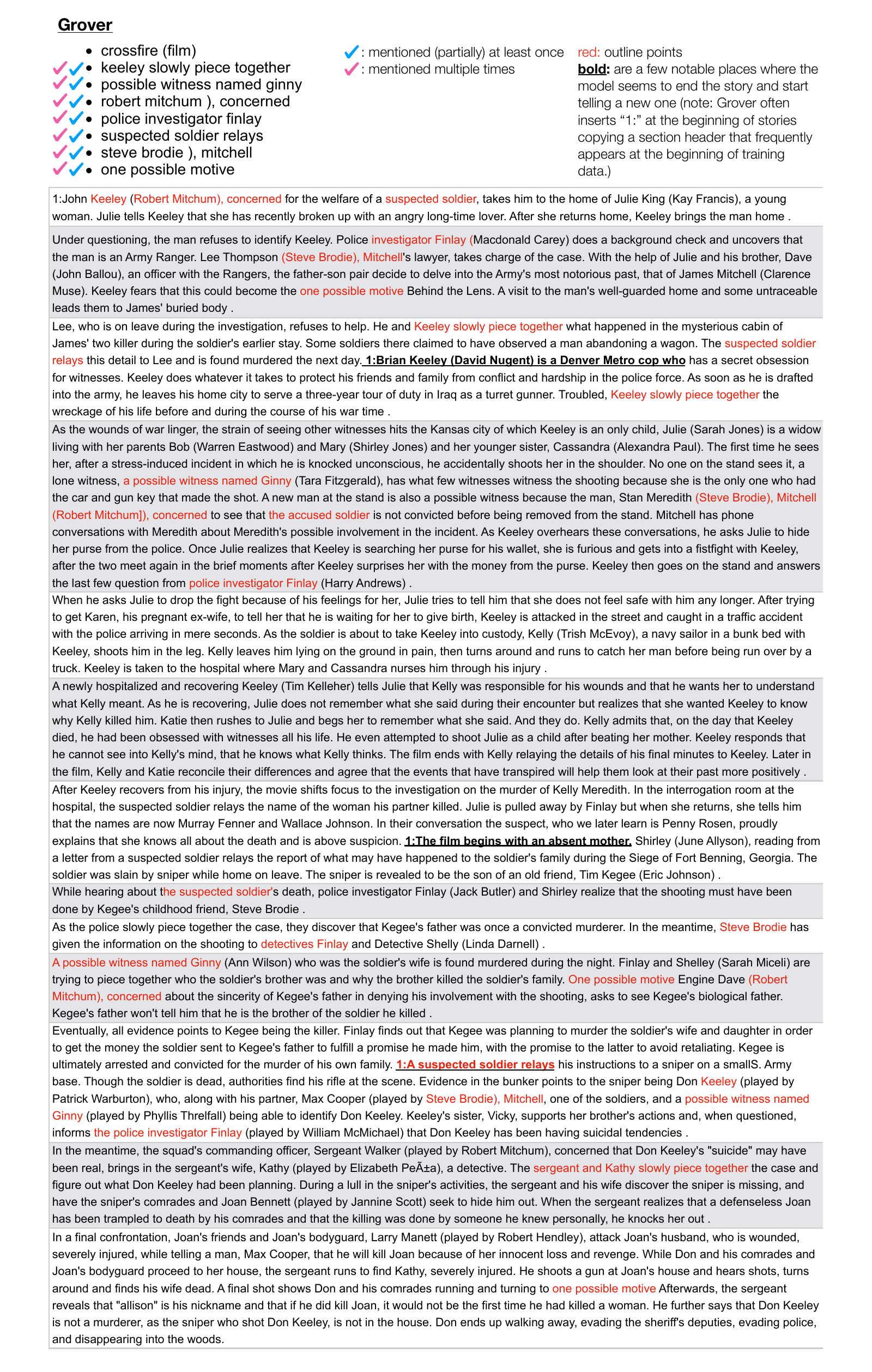}
    \caption{Example document generated using Grover}
    \label{fig:gen:grover}
\end{figure*}

\begin{figure*}
\centering
\includegraphics[page=4,trim={0 17in 0 0.75in}, clip,width=\textwidth]{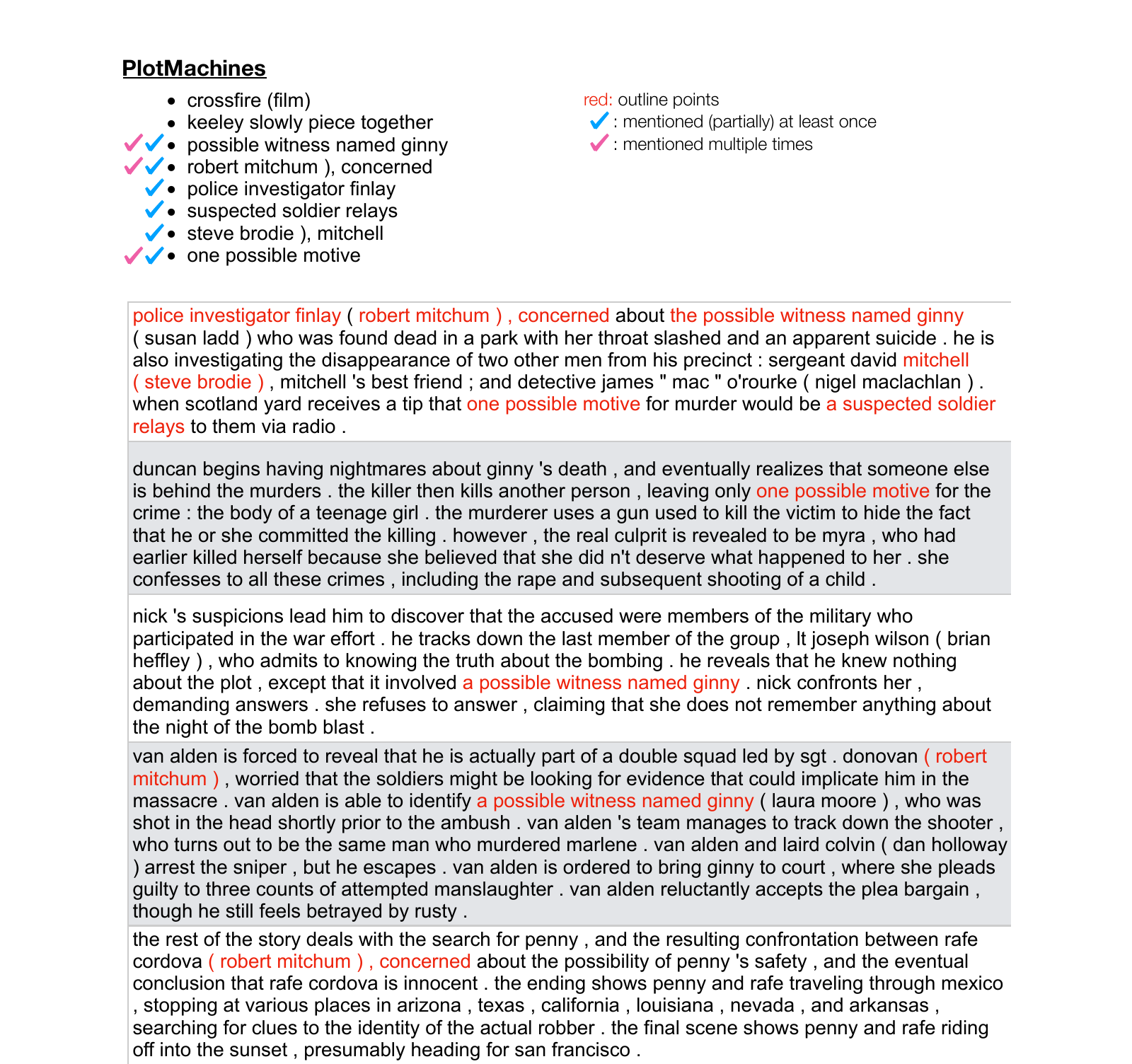}
    \caption{Example document generated using \modelname{}}
    \label{fig:gen:pm1}
\end{figure*}

\begin{figure*}
\centering
\includegraphics[page=1,trim={0 14in 0 2.in}, clip,width=\textwidth]{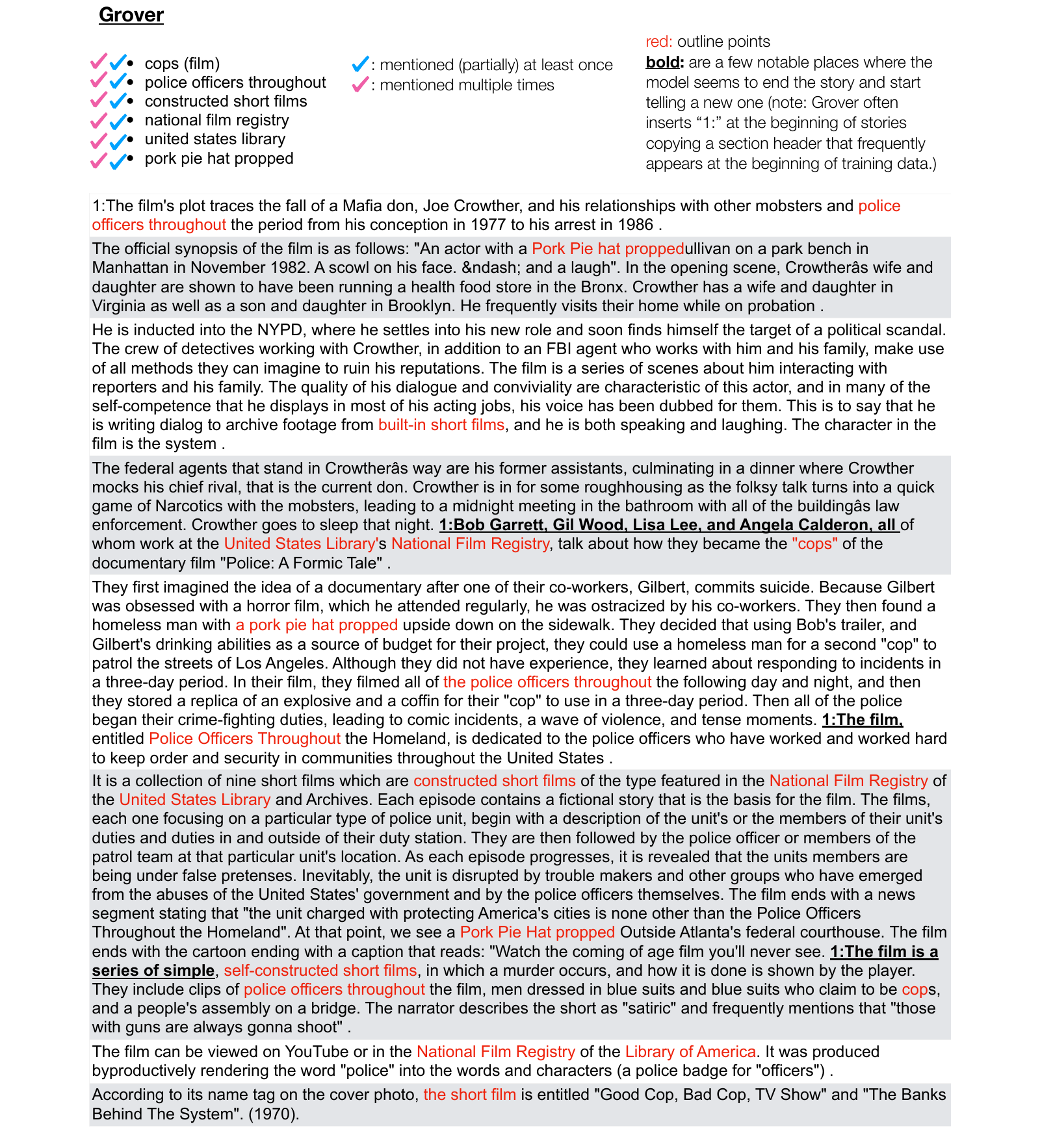}
    \caption{Example document generated using \grover{}}
    \label{fig:gen:grover2}
\end{figure*}

\begin{figure*}
\centering
\includegraphics[page=2,trim={0 18in 0 2.in}, clip,width=\textwidth]{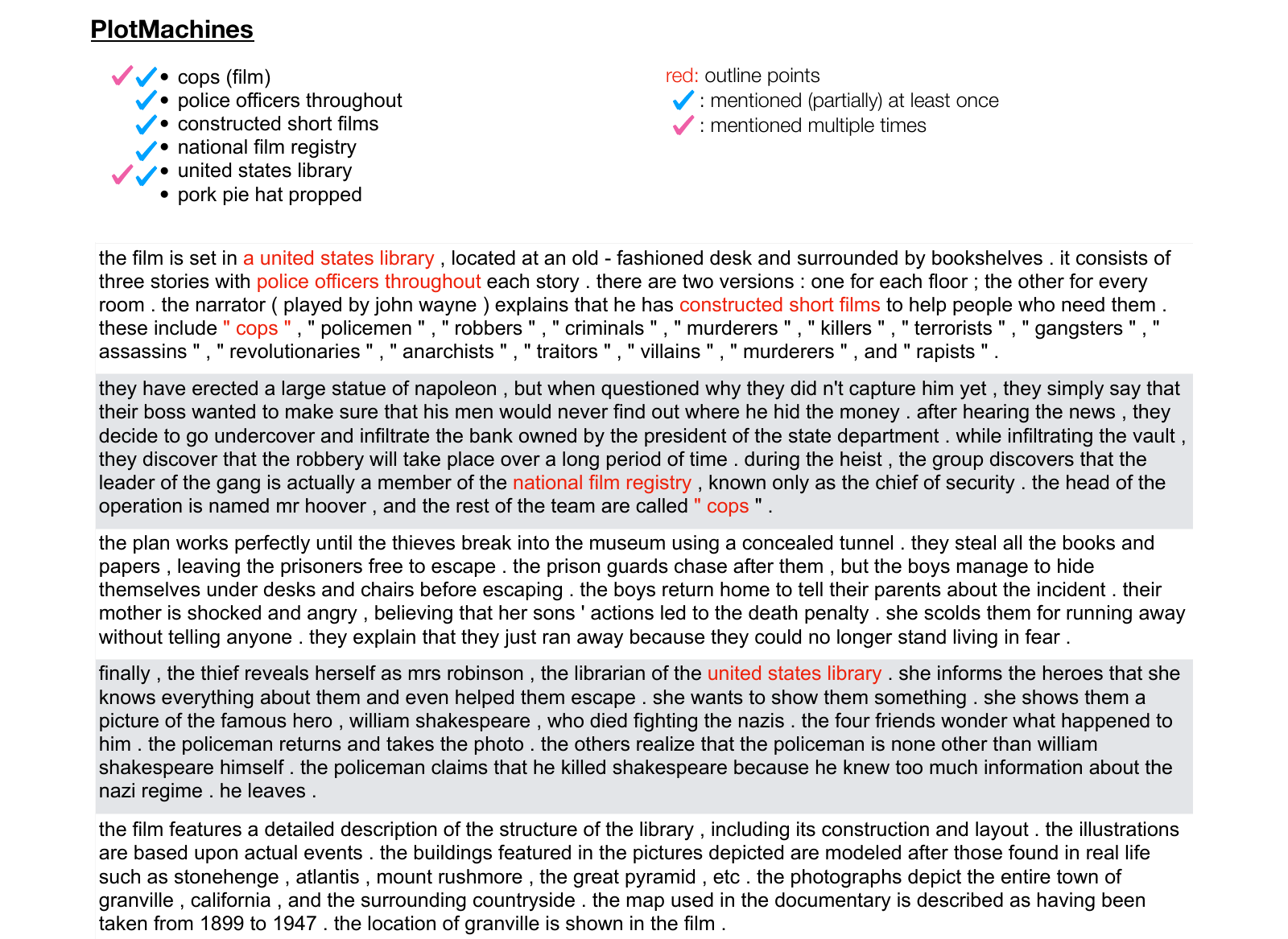}
    \caption{Example document generated using \modelname{}}
    \label{fig:gen:pm2}
\end{figure*}

\begin{figure*}
\centering
\includegraphics[page=5,trim={0 19.5in 0 2.4in}, clip,width=\textwidth]{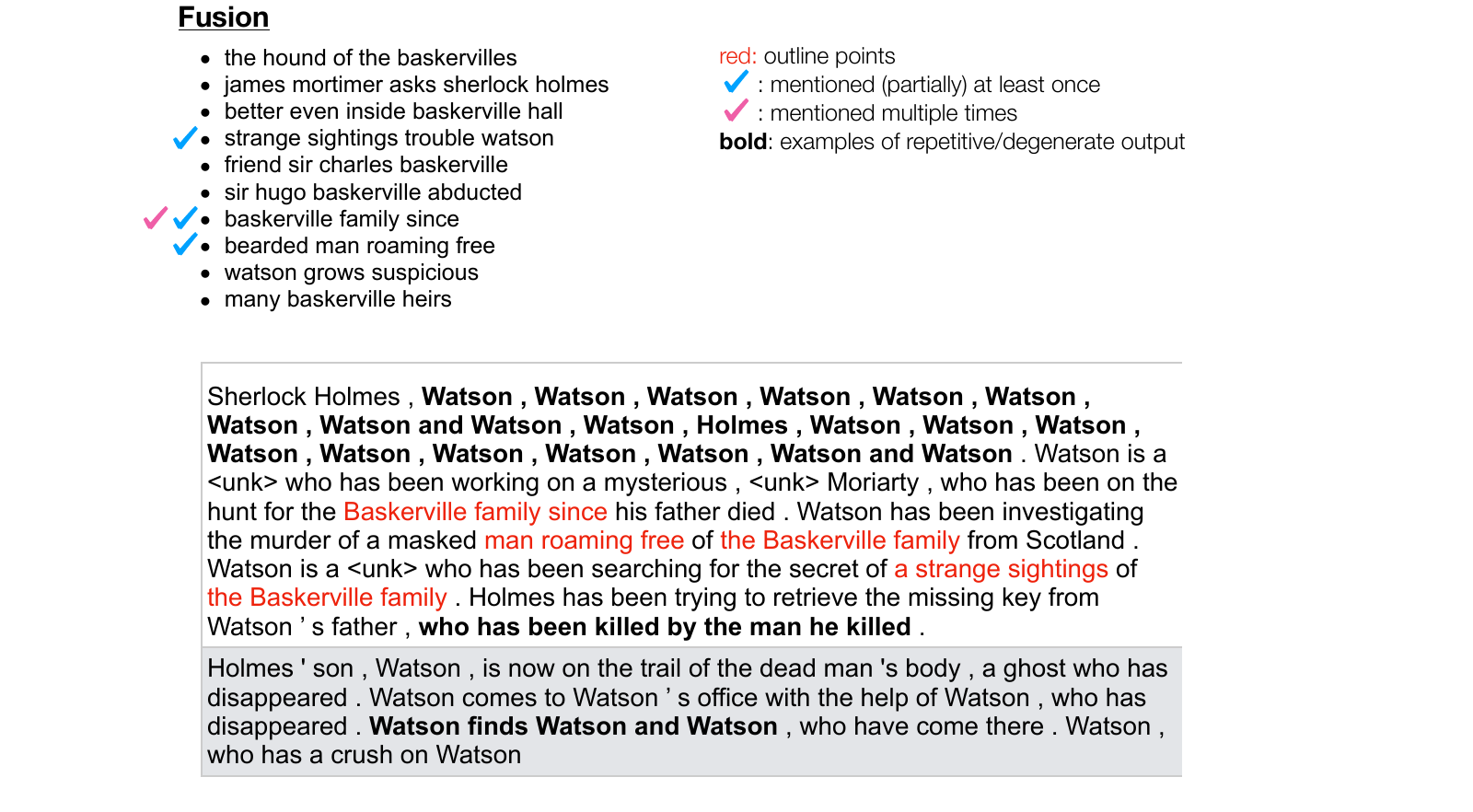}
    \caption{Example document generated using the Fusion model \citep{WritingPrompts}}
    \label{fig:gen:fusion}
\end{figure*}

\begin{figure*}
\centering
\includegraphics[page=6,trim={0 16in 0 2.4in}, clip,width=\textwidth]{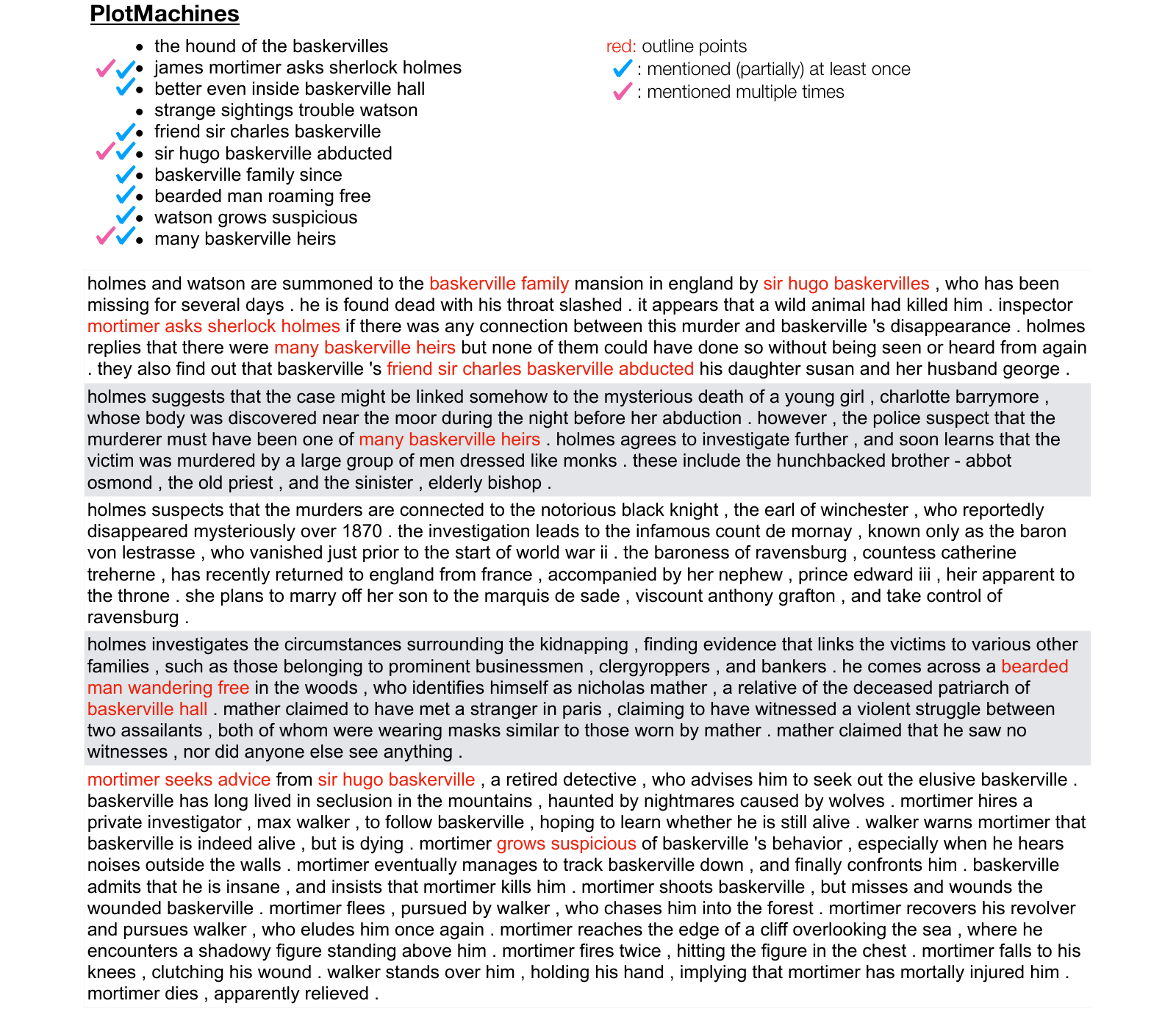}
    \caption{Example document generated using \modelname{}}
    \label{fig:gen:pm3}
\end{figure*}

We show example full stories from the Wikiplots validation set comparing outputs from:
\begin{itemize}
    \item \grover{} (Figure~\ref{fig:gen:grover}) and \modelname{} (Figure~\ref{fig:gen:pm1})
    \item \grover{} (Figure~\ref{fig:gen:grover2}) and \modelname{} (Figure~\ref{fig:gen:pm2}) 
    \item Fusion \cite{WritingPrompts} (Figure~\ref{fig:gen:fusion}) and \modelname{} (Figure~\ref{fig:gen:pm3}) 
\end{itemize}

In the examples, we highlight outline points that are mentioned in red.  We also bold a few sections in the \grover{} output where the model notably ends the story and starts a new one.
Examples indicate that \grover{} often finishes the story and then starts a new story partway through the document.  This shortcoming may help explain why \grover{} over-repeats outline points and why humans judge it to be more repetitive and less consistently relevant. 
In contrast, our models adhere more to a beginning-middle-ending structure.

We also show additional examples of introduction and conclusion paragraphs generated by \modelname{}  (Table~\ref{tab:introsandconcl}), demonstrating the discourse the model has learned.  For example, the model often starts stories by setting the scene (e.g. ``In the early 1950s, a nuclear weapons testing continues ....'') and often ends with a definitive closing action (e.g. ``... the film ends with humperdinck and buttercup riding off into the sunset.'')

\begin{table*}[]
    \centering
    \footnotesize
    \begin{tabular}{|cp{13cm}|}
    \hline
    Paragraph type & Paragraph \\
    \hline
         intro & in the early 1950s, a nuclear weapons testing continues at an underwater hydrogen bomb test site. scientists are concerned that it may be too dangerous to detonate without being detected by radar and radiation detectors. government sends paleontologist kyohei yamane ( kim kap ) to investigate. he is killed when his boat explodes while on shore patrol. as evidence describes damage consistent with sabotage of oil rigs, they conclude there must have been more than one way inside the facility. meanwhile, military research has discovered a deep underwater natural habitat alongside others where water can not be mined for life - saving purposes. \\
         \hline
         intro & the novel is set in a post - apocalyptic future where earth almost uninhabitable, with only one habitable planet for habitation and an intersolar system police force ( rf ) to maintain order. the story begins when " cowboy bebop ", who has been living on his homeworld of nepal since he was 12 years old, returns from space after being stranded by a comet that destroyed most of the interstellar civilization. he finds himself at home as well as friends among other characters. \\
         \hline
         intro & in 1933, joker terrorizes gotham city by murdering the mayor of new york. commissioner gordon is called to defend gotham whenever crime strikes are occurring and he has his own reasons for doing so : a corrupt police lieutenant eckhardt ( james stewart ) wants napier captured alive ; an elderly woman who was once part of batman 's gang tries to kill him but instead accidentally drops her gun into the water. joker also becomes obsessed with capturing the joker. meanwhile, photojournalist vicki vale begin their investigation on batman as well as other characters from the newspaper " big daddy " and " the joker ". \\
         \hline
         conclusion &  humperdinck arranges for buttercup to get married to a powerful don juan carlos, who is rumored to be able to control the entire province. humperdinck secretly orders rugen to kidnap buttercup and bring her to him. rugen succeeds in kidnapping buttercup, but humperdinck kidnaps her anyway. buttercup manages to free herself and flee with humperdinck, but is captured by manuela, who accuses humperdinck of trying to keep her prisoner. humperdinck swears revenge on manuela and his henchmen, and rescues buttercup just in time. the pair head north to santa fe, where humperdinck uses his magic powers to heal buttercup 's wounds. the couple settle in a small cabin owned by mrs mccluskey, who introduces buttercup to mr smith, a blacksmith. humperdinck 's plan backfires when mr smith is attacked by apache indians, and humperdinck saves him. the film ends with humperdinck and buttercup riding off into the sunset .\\
         \hline
         conclusion  & stevens and angel eyes sneak into the church hall and steal a bible. stevens opens the book and reads passages from psalms 118 to 350 bc. stevens closes the book and hands it to angel eyes. angel eyes then places stevens ' hand atop the cross and prepares to strike. stevens grabs hold of angel eyes and begs him to reconsider. stevens pleads with angel eyes to listen to reason. angel eyes makes stevens tell him why he left the confederacy. stevens tells him that he was betrayed by his mother and sister and that he needs redemption. stevens then lies and tells angel eyes that he ca n't forgive him. stevens then walks away. angel eyes watches him disappear into the night .\\
         \hline
         {conclusion} & in 1987, toscani meets harrison at a bar, where harrison confesses that he orchestrated the bombing of harrison 's hotel room. harrison promised justice for his friends and family, but toscani refused to believe him. harrison pleads with toscani to let him live, but toscani rejects him. toscani drives away, and harrison follows him. toscani breaks down crying, realizing that he has failed. harrison promises justice for his victims, and toscani smiles sadly .\\
         \hline
    \end{tabular}
    \caption{Example introduction and conclusion paragraph generations from   \modelname{} using the Wikiplots validation set.}
    \label{tab:introsandconcl}
\end{table*}

\end{document}